%% file: icml2020_conference.tex
\documentclass{article}
\usepackage[final]{neurips_2020}

\usepackage[utf8]{inputenc} 
\usepackage[T1]{fontenc}    
\usepackage{hyperref}       
\usepackage{url}            
\usepackage{booktabs}       
\usepackage{amsfonts}       
\usepackage{nicefrac}       
\usepackage{microtype}      

\input{packages.tex}
\input{macros.tex}

\input{math_commands.tex}

\title{Compositional Visual Generation with Energy Based Models}

\author{%
  Yilun Du \\
  MIT CSAIL\\
  yilundu@mit.edu\\
  \And
  Shuang Li \\
  MIT CSAIL\\
  lishuang@mit.edu \\
  \And
  Igor Mordatch \\
  Google Brain\\
  imordatch@google.com\\
  \And
}

\begin{document}
\maketitle

\input{text/abstract.tex}
\input{text/introduction.tex}

\input{text/related_work.tex}
\input{text/method.tex}
\input{text/evaluation.tex}

\input{text/discussion.tex}
\input{text/acknowledgement.tex}
\input{text/broader_impacts.tex}
\newpage

{\small
\bibliographystyle{plainnat}
\bibliography{icml2020_conference}
}

\appendix
\renewcommand{\thesection}{A.\arabic{section}}
\renewcommand{\thefigure}{A\arabic{figure}}
\setcounter{section}{0}
\setcounter{figure}{0}
\input{text/appendix}

\end{document}

%% file: packages.tex
\usepackage{color,xcolor}
\usepackage{epsfig}
\usepackage{graphicx}

\usepackage{adjustbox}
\usepackage{array}
\usepackage{booktabs}
\usepackage{colortbl}
\usepackage{float,wrapfig}
\usepackage{hhline}
\usepackage{multirow}
\usepackage{subcaption} 

\usepackage{amsmath,amsfonts,amsthm,amssymb}
\usepackage{bm}
\usepackage{nicefrac}
\usepackage{microtype}

\usepackage{changepage}
\usepackage{extramarks}
\usepackage{fancyhdr}
\usepackage{lastpage}
\usepackage{setspace}
\usepackage{soul}
\usepackage{xspace}

\usepackage{url}

\usepackage{algorithm, algorithmic}
\usepackage{enumerate}
\usepackage{todonotes} 

\usepackage{titlesec}
\usepackage{makecell}

\usepackage{amsmath,amsthm,amssymb}
\usepackage{mathtools}

%% file: macros.tex
\newcolumntype{L}[1]{>{\raggedright\let\newline\\\arraybackslash\hspace{0pt}}m{#1}}
\newcolumntype{C}[1]{>{\centering\let\newline\\\arraybackslash\hspace{0pt}}m{#1}}
\newcolumntype{R}[1]{>{\raggedleft\let\newline\\\arraybackslash\hspace{0pt}}m{#1}}

\newcommand{\eqn}[1]{Equation~\ref{#1}}
\newcommand{\fig}[1]{Figure~\ref{#1}}
\newcommand{\tbl}[1]{Table~\ref{#1}}

\newcommand{\ignore}[1]{}

\makeatletter
\DeclareRobustCommand\onedot{\futurelet\@let@token\@onedot}
\def\@onedot{\ifx\@let@token.\else.\null\fi\xspace}

\makeatother

\definecolor{MyDarkBlue}{rgb}{0,0.08,1}
\definecolor{MyDarkGreen}{rgb}{0.02,0.6,0.02}
\definecolor{MyDarkRed}{rgb}{0.8,0.02,0.02}
\definecolor{MyDarkOrange}{rgb}{0.40,0.2,0.02}
\definecolor{MyPurple}{RGB}{111,0,255}
\definecolor{MyRed}{rgb}{1.0,0.0,0.0}
\definecolor{MyGold}{rgb}{0.75,0.6,0.12}
\definecolor{MyDarkgray}{rgb}{0.66, 0.66, 0.66}
\definecolor{MyDarkPink}{rgb}{1,0.078,0.57}

\newcommand{\myparagraph}[1]{\vspace{-8pt}\paragraph{#1}}

\titlespacing*{\section}{0pt}{0pt plus 2pt minus 2pt}{0pt plus 2pt minus 2pt}
\titlespacing\subsection{0pt}{0pt plus 1pt minus 1pt}{0pt plus 1pt minus 1pt}

\DeclarePairedDelimiterX{\infdivx}[2]{(}{)}{%
  #1\;\delimsize|\delimsize|\;#2%
}

\def\x{\mathbf{x}}
\def\sx{\tilde{\x}}

%% file: math_commands.tex
\usepackage{amsmath,amsfonts,bm}

\def\eqref#1{equation~\ref{#1}}

\def\1{\bm{1}}

\DeclareMathAlphabet{\mathsfit}{\encodingdefault}{\sfdefault}{m}{sl}
\SetMathAlphabet{\mathsfit}{bold}{\encodingdefault}{\sfdefault}{bx}{n}



%% file: text/abstract.tex
\begin{abstract}


A vital aspect of human intelligence is the ability to compose increasingly complex concepts out of simpler ideas, enabling both rapid learning and adaptation of knowledge. In this paper we show that energy-based models can exhibit this ability by directly combining probability distributions. Samples from the combined distribution correspond to compositions of concepts. For example, given one distribution for smiling face images, and another for male faces, we can combine them to generate smiling male faces. This allows us to generate natural images that simultaneously satisfy conjunctions, disjunctions, and negations of concepts. We evaluate compositional generation abilities of our model on the CelebA dataset of natural faces and synthetic 3D scene images. We showcase the breadth of unique capabilities of our model, such as the ability to continually learn and incorporate new concepts, or infer compositions of concept properties underlying an image. 


\end{abstract}

%% file: text/introduction.tex
\section{Introduction}

Humans are able to rapidly learn new concepts and continuously integrate them among prior knowledge. The core component in enabling this is the ability to compose increasingly complex concepts out of simpler ones as well as recombining and reusing concepts in novel ways \citep{fodor2002compositionality}. By combining a finite number of primitive components, humans can create an exponential number of new concepts, and use them to rapidly explain current and past experiences \citep{lake2017building}. We are interested in enabling such capabilities in machine learning systems, particularly in the context of generative modeling. 

Past efforts have attempted to enable compositionality in several ways. One approach decomposes data into disentangled factors of variation and situate each datapoint in the resulting - typically continuous - factor vector space \citep{vedantam2017generative,higgins2017scan}. The factors can either be explicitly provided or learned in an unsupervised manner. In both cases, however, the dimensionality of the factor vector space is fixed and defined prior to training. This makes it difficult to introduce new factors of variation, which may be necessary to explain new data, or to taxonomize past data in new ways. Another approach to incorporate the compositionality is to spatially decompose an image into a collection of objects, each object slot occupying some pixels of the image defined by a segmentation mask \citep{van2018case,greff2019multi}. Such approaches can generate visual scenes with multiple objects, but may have difficulty in generating interactions between objects. These two incorporations of compositionality are considered distinct, with very different underlying implementations.

In this work\footnote{Code and data available at \url{https://energy-based-model.github.io/compositional-generation-inference/}}, we propose to implement the compositionality via energy based models (EBMs). Instead of an explicit vector of factors that is input to a generator function, or object slots that are blended to form an image, our unified treatment defines factors of variation and object slots via energy functions. Each factor is represented by an individual scalar energy function that takes as input an image and outputs a low energy value if the factor is exhibited in the image. Images that exhibit the factor can then be generated implicitly through an Markov Chain Monte Carlo (MCMC) sampling process that minimizes the energy. Importantly, it is also possible to run MCMC process on some \emph{combination} of energy functions to generate images that exhibit multiple factors or multiple objects, in a globally coherent manner.

There are several ways to combine energy functions. One can add or multiply distributions as in mixtures \citep{shazeer2017outrageously,greff2019multi} or products \citep{hinton2002training} of experts. We view these as probabilistic instances of logical operators over concepts. Instead of using only one, we consider three operators: logical conjunction, disjunction, and negation (illustrated in Figure \ref{fig:comp_cartoon}). We can then flexibly and recursively combine multiple energy functions via these operators. More complex operators (such as implication) can be formed out of our base operators.

\input{figText/comp_cartoon.tex}

EBMs with such composition operations enable a breadth of new capabilities - among them is a unique approach to continual learning. Our formulation defines concepts or factors implicitly via examples, rather than pre-declaring an explicit latent space ahead of time. For example, we can create an EBM for concept "black hair" from a dataset of face images that share this concept. New concepts (or factors), such as hair color can be learned by simply adding a new energy function and can then be combined with energies for previously trained concepts. This process can repeat continually. This view of few-shot concept learning and generation is similar to work of \citep{reed2017few}, with the distinction that instead of learning to generate holistic images from few examples, we learn \emph{factors} from examples, which can be composed with other factors. A related advantage is that finely controllable image generation can be achieved by specifying the desired image via a collection of logical clauses, with applications to neural scene rendering \citep{eslami2018neural}.

Our contributions are as follows: first, while composition of energy-based models has been proposed in abstract settings before \citep{hinton2002training}, we show that it can be used to generate plausible natural images. Second, we propose a principled approach to combine independent trained energy models based on logical operators which can be chained recursively, allowing controllable generation based on a collection of logical clauses at test time. Third, by being able to recursively combine independent models,  we show our approach allows us to extrapolate to new concept combinations, continually incorporate new visual concepts for generation, and infer concept properties compositionally.








%% file: figText/comp_cartoon.tex
\begin{figure*}[t]
\begin{center}
\includegraphics[width=0.9\textwidth]{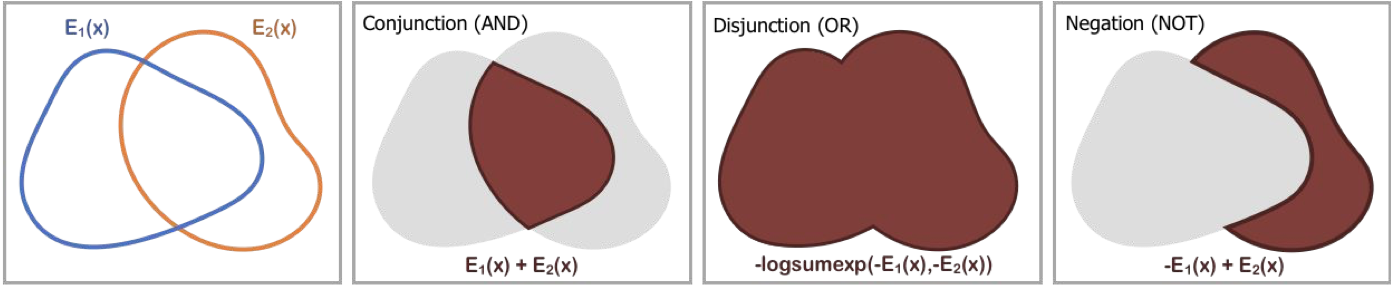}
\end{center}
\vspace{-10pt}
\caption{\small Illustration of logical composition operators over energy functions $E_1$ and $E_2$ (drawn as level sets where red = valid areas of samples, grey = invalid areas of samples). 
}
\label{fig:comp_cartoon}
\vspace{-10pt}
\end{figure*}

%% file: text/related_work.tex
\section{Related Work}
Our work draws on results in energy based models - see \citep{lecun2006tutorial} for a comprehensive review. A number of methods have been used for inference and sampling in EBMs, from Gibbs Sampling \citep{hinton2006fast}, Langevin Dynamics \citep{xie2016theory, du2019implicit}, Path Integral methods \citep{du2019plan} and learned samplers \citep{kim2016deep, song2018learning}. In this work, we apply EBMs to the task of compositional generation.


Compositionality has been incorporated in representation learning (see \citep{andreas2019measuring} for a summary) and generative modeling. One approach to compositionality has focused on learning disentangled factors of variation \citep{Higgins2017Beta, Kulkarni2015Deep, vedantam2017generative}. Such an approach allows for the combination of existing factors, but does not allow the addition of new factors.
A different approach to compositionality includes learning various different pixel/segmentation masks for each concept \citep{greff2019multi, Gregor2015DRAW}. However such a factorization may have difficulty capturing the global structure of an image, and in many cases different concepts cannot be explicitly factored using attention masks. Individual compositions of factors can also be seen as a domain translation problem \citep{Benaim2019DomainIntersectionDifference, press2018emerging, abs-1906.06558}. Such an formulation, however, requires separate retraining of models for each considered composition.

In contrast, our approach towards compositionality focuses on composing separate learned probability distribution of concepts. Such an approach allows viewing factors of variation as constraints \citep{mnih2005learning}. In prior work, \citep{hinton1999products} show that products of EBMs can be used to decompose complex generative modeling problems to simpler ones. \citep{vedantam2017generative} further apply products of distributions over the latent space of VAE to define compositions. \citep{higgins2017scan} show that additional compositions in VAE latent space. Both of them rely on joint training to learn compositions of a fixed number of concepts. In contrast, in this work, we show how we can realize concept compositions using completely \textbf{independently} trained probability distributions. Furthermore, we introduce three compositional logical operators of conjunction, disjunction and negation can be realized and nested together through manipulation of independent probability distributions of each concept.  

Our compositional approach is inspired by the goal of continual lifelong learning - see \citep{parisi} for a thorough review. New concepts can be composed with past concepts by combining new independent probability distributions. Many methods in continual learning are focused on how to overcome catashtophic forgetting \citep{kirkpatrick2017overcoming,li2017learning}, but do not support dynamically growing capacity. Progressive growing of the models \citep{rusu2016progressive} has been considered, but is implemented at the level of the model architecture, whereas our method composes independent models together. 



%% file: text/method.tex
\section{Method}



In this section, we first give an overview of the Energy-Based Model formulation we use and introduce three logical operators over these models. We then discuss the unique properties such a form of compositionality enables.

\subsection{Energy Based Models}
EBMs represent data by learning an unnormalized probability distribution across the data. For each data point $\x$, an energy function $E_\theta(\x)$, parameterized by a neural network, outputs a scalar real energy such that the model distribution  
\vspace{-5pt}
\begin{align}
p_\theta(x) & \propto e^{-E_\theta(x)}.
\label{eq:prob}
\end{align}
To train an EBM on a data distribution $p_D$, we use contrastive divergence \citep{hinton1999products}. In particular we use the methodology defined in \citep{du2019implicit}, where a Monte Carlo estimate (\eqn{eq:mce}) of maximum likelihood $\mathcal{L}$ is minimized with the following gradient
\begin{align}
\nabla_\theta \mathcal{L} = \mathbb{E}_{x^+ \sim p_D}{ \nabla_\theta E_\theta(x^+)} - \mathbb{E}_{x^- \sim p_\theta}{ \nabla_\theta E_\theta(x^-)}.
\label{eq:mce}
\end{align}
To sample $x^-$ from $p_\theta$ for both training and generation, we use MCMC based off Langevin dynamics \citep{welling2011bayesian}. Samples are initialized from uniform random noise and are iteratively refined using
\vspace{-5pt}
\begin{align}
\sx^k &= \sx^{k-1} - \frac{\lambda}{2} \nabla_\x E_\theta (\sx^{k-1}) + \omega^k, \; \omega^k \sim \mathcal{N}(0,\lambda),
\label{eq:langevin}
\end{align}
where $k$ is the $k^{th}$ iteration step and $\lambda$ is the step size. We refer to each iteration of Langevin dynamics as a negative sampling step. We note that this form of sampling allows us to use the gradient of the combined distribution to generate samples from distributions composed of $p_\theta$ and the other distributions. We use this ability to generate from multiple different compositions of distributions. To enable high resolution image generation, we further apply techniques for improving EBM training described in \citep{du2020cd} on CelebA images.

\input{figText/venn.tex}
\subsection{Composition of Energy-Based Models}
\label{sec:composition}
We next present different ways that EBMs can compose. We consider a set of independently trained EBMs, $E(\x|c_1), E(\x|c_2), \ldots,  E(\x|c_n)$, which are learned conditional distributions on underlying concept codes $c_i$. Latent codes we consider include position, size, color, gender, hair style, and age, which we also refer to as concepts. Figure \ref{fig:venn} shows three concepts and their combinations on the CelebA face dataset and attributes.

\myparagraph{Concept Conjunction} In concept conjunction, given separate independent concepts (such as a particular gender, hair style, or facial expression), we wish to construct an output with the specified gender, hair style, and facial expression -- the combination of each concept. Since the likelihood of an output given a set of specific concepts is equal to the product of the likelihood of each individual concept, we have \eqn{eq:joint}, which is also known as the product of experts \citep{hinton2002training}:
\vspace{-2pt}
\begin{align}
p(x|c_1 \; \text{and} \; c_2, \ldots, \; \text{and} \; c_i) = \prod_i p(x|c_i) \propto e^{-\sum_i E(x|c_i)}.
\label{eq:joint}
\end{align}
We can thus apply \eqn{eq:langevin} to the distribution that is the sum of the energies of each concept.  We sample from this distribution using \eqn{eq:langevin_comb} to sample from the joint concept space with $\omega^k \sim \mathcal{N}(0,\lambda)$.
\vspace{-2pt}
\begin{equation}
\sx^k = \sx^{k-1} - \frac{\lambda}{2} \nabla_\x \sum_i E_\theta (\sx^{k-1}|c_i) + \omega^k.
\label{eq:langevin_comb}
\vspace{-10pt}
\end{equation}
\myparagraph{Concept Disjunction} In concept disjunction, given separate concepts such as the colors red and blue, we wish to construct an output that is either red or blue. This requires a distribution that has probability mass when any chosen concept is true. A natural choice of such a distribution is the sum of the likelihood of each concept:
\vspace{-2pt}
\begin{align}
p(x|c_1 \; \text{or} \; c_2, \ldots \; \text{or}  \; c_i) \propto \sum_i p(x|c_i) / Z(c_i).
\label{eq:or}
\end{align}
where $Z(c_i)$ denotes the partition function for each concept. A tractable simplification becomes available if we assume all partition functions $Z(c_i)$ to be equal
\vspace{-2pt}
\begin{align}
\sum_i p(x|c_i) \propto \sum_i e^{-E(x|c_i)} = e^{ \text{logsumexp}(-E(x|c_1), -E(x|c_2), \ldots, -E(x|c_i))},
\label{eq:or_simple}
\end{align}

where $\text{logsumexp}(f_1, \ldots ,f_N) = \log \sum_i \exp(f_i)$. We can thus apply \eqn{eq:langevin} to the distribution that is a negative smooth minimum of the energies of each concept to obtain \eqn{eq:langevin_add} to sample from the disjunction concept space:
\vspace{-2pt}
\begin{align}
& \sx^k = \sx^{k-1} - \frac{\lambda}{2} \nabla_\x \text{logsumexp}(-E(x|c_1), -E(x|c_2), \ldots, -E(x|c_i))  + \omega^k,
\label{eq:langevin_add}
\end{align}
where $\omega^k \sim \mathcal{N}(0,\lambda)$.
While the assumption that leads to Equation \ref{eq:or_simple} is not guaranteed to hold in general, in our experiments we empirically found the partition function $Z(c_i)$ estimates to be similar across  partition functions (see Appendix) and also analyze cases in which partitions functions are different in the Appendix. Furthermore, the resulting generation results do exhibit equal distribution across disjunction constituents in practice as seen in \tbl{tbl:disjunction_tbl}.

\myparagraph{Concept Negation} In concept negation, we wish to generate an output that does not contain the concept. Given a color red, we want an output that is of a different color, such as blue. Thus, we want to construct a distribution that places high likelihood to data that is outside a given concept. One choice is a distribution inversely proportional to the concept. Importantly, negation must be defined with respect to another concept to be useful. The opposite of alive may be dead, but not inanimate. Negation without a data distribution is not integrable and leads to a generation of chaotic textures which, while satisfying absence of a concept, is not desirable. Thus in our experiments with negation we combine it with another concept to ground the negation and obtain an integrable distribution:
\vspace{-2pt}
\begin{align}
p(x| \text{not}(c_1), c_2) \propto \frac{p(x|c_2)}{p(x|c_1)^\alpha} \propto e^{ \alpha  E(x|c_1) - E(x|c_2) }.
\label{eq:neg}
\end{align}
%
We found the smoothing parameter $\alpha$ to be a useful regularizer (when $\alpha = 0$ we arrive at uniform distribution) and we use $\alpha = 0.01$ in our experiments. The above equation allows us to apply Langevin dynamics to obtain \eqn{eq:langevin_neg} to sample concept negations.
\vspace{-2pt}
\begin{align}
& \sx^k = \sx^{k-1} - \frac{\lambda}{2} \nabla_\x (\alpha E(x|c_1) - E(x|c_2)) + \omega^k, 
\label{eq:langevin_neg}
\end{align}
where $\omega^k \sim \mathcal{N}(0,\lambda)$.

\myparagraph{Recursive Concept Combinations} We have defined the three classical symbolic operators for concept combinations. These symbolic operators can further be recursively chained on top of each to specify more complex logical operators at test time. To our knowledge, our approach is the only approach enabling such compositionality across independently trained models.

%% file: figText/venn.tex
\begin{figure*}[t]
\begin{center}
\includegraphics[width=0.8\textwidth]{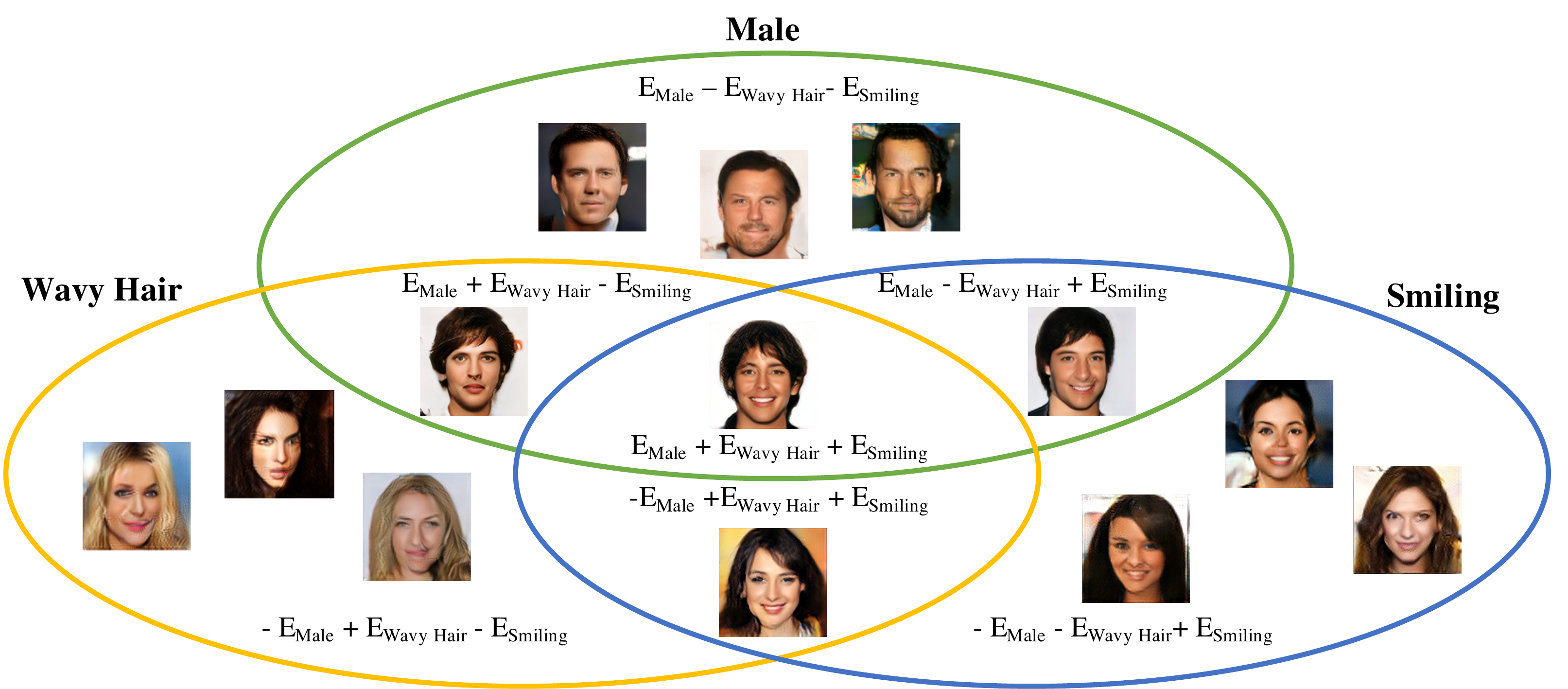}
\end{center}
\vspace{-10pt}
\caption{\small Concept conjunction and negation. All the images are generated through the conjunction and negation of energy functions. For example, the image in the central part is the conjunction of male, black hair, and smiling energy functions. Equations for composition explained in page 4.}
\label{fig:venn}
\end{figure*}

%% file: text/evaluation.tex
\section{Experiments}
We perform empirical studies to answer the following questions: (1) Can EBMs exhibit concept compositionality (such as concept negation, conjunction, and disjunction) in generating images? 
(2) Can we take advantage of concept combinations to learn new concepts in a continual manner? 
(3) Does explicit factor decomposition enable generalization to novel combinations of factors?
(4) Can we perform concept inference across multiple inputs?

In the appendix, we further show that approach enables better generalization to novel combinations of factors by learning explicit factor decompositions.

\input{figText/celeba_mujoco_combine.tex}
\subsection{Setup}
We perform experiments on 64x64 object scenes rendered in MuJoCo \citep{Todorov2012MuJoCo} (MuJoCo Scenes) and the 128x128 CelebA dataset. For MuJoCo Scene images, we generate a central object of shape either sphere, cylinder, or box of varying size and color at different positions, with some number of (specified) additional background objects. Images are generated with varying lighting and objects.

We use the ImageNet32x32 architecture and ImageNet128x128 architecture from \citep{du2019implicit} with the Swish activation \citep{ramachandran2017searching} on MuJoCo and CelebA datasets. Models are trained on MuJoCo datasets for up to 1 day on 1 GPU and for 1 day on 8 GPUs for CelebA. More training details and model architecture can be found in the appendix.

\input{figText/disjunction_conjunction.tex}

\subsection{Compositional Generation}
\noindent \textbf{Quantitative evaluation.} 
We first evaluate compositionality operations of EBMs in Section~\ref{sec:composition}.
%
To quantitatively evaluate generation, we use the MuJoCo Scenes dataset. We train a supervised classifier to predict the object position and color on the MuJoCo Scenes dataset. Our classifier obtains $99.3\%$ accuracy for position and $99.9\%$ for color on the test set. We also train seperate conditional EBMs on the concepts of position and color. 
For a given positional generation then, if the predicted position (obtained from a supervised classifier on generated images) and original conditioned generation position is smaller than 0.4, then a generation is consider correct. A color generation is correct if the predicted color is the same as the conditioned generation color.


In \tbl{tbl:disjunction_tbl}, we quantitatively evaluate the quality of generated images given combinations of conjunction, disjunction, and negation on the color and position concepts. When using either Color or Position EBMs, the respective accuracy is high. Conjunction(Position, Color) has high position and color accuracies which demonstrates that an EBM can combine different concepts. Under Conjunction(Position, Negation(Color)), the color accuracy drops to below that of Color EBM. This means negating a concept reduces the likelihood of the concept. The same conclusion for Conjunction(Negation(Position), Color). We compare with the approach in \citep{vedantam2017generative}, using the author's online github repo, and find it produces blurrier and worse results.


To evaluate disjunction, we set Position 1 to be a random point in the bottom left corner of a grid and Position 2 to be a random point in the top right corner of a grid. 
The average results over 1000 generated images are reported in Table~\ref{tbl:disjunction_tbl}. Position 1 EBM or Position 2 EBM can obtain high accuracy in predicting their own positions. Disjunction(Position 1, Position 2) EBM generate images that are roughly evenly distributed between Position 1 and Position 2, indicating the disjunction can combine concepts additively. This trend further holds with conjunction, with 
Disjunction(Conjunction(Position 1, Color 1),Conjunction(Position 2, Color 2)) also being evenly distributed.

We further investigate implication using a composition of conjunctions and negations in EBMs. We consider the term (Position 1 AND (NOT Color 1)) AND ... AND (Position 1 AND (NOT Color 4)), which implicates Color 5. We find that are generations obtain 0.982 accuracy for Color 5.
\myparagraph{Qualitative evaluation.} 
We further provide qualitative visualizations of conjunction, disjunction, and negation operations on both MuJoCo Scenes and CelebA datasets.

\textit{Concept Conjunction:} In \fig{fig:celeba_combine}, we show the conjunction of EBMs is able to combine multiple independent concepts, such as age, gender, smile, and wavy hair, and get more precise generations with each energy models. Our composed generations  obtain a FID of 45.3, compared to an FID of 64.5 of an SNGAN model trained on data conditioned on all four attributes. Our generations are also significantly more diverse than that of GAN model (average pixel MSE of 64.5 compared to 55.4 of the GAN model). Similarily, EBMs can combine independent concepts of shape, position, size, and color to get more precise generations in \fig{fig:mujoco_combine}. We also show results of conjunction with other logical operators in \fig{fig:andor}.

\textit{Concept Negation:} In \fig{fig:andor}, row 4 shows images that are opposite to the trained concept using negation operation. Since concept negation operation should accompany with another concept as described in Section~\ref{sec:composition}, we use ``smiling`` as the second concept. The images in row 4 shows the negation of male AND smiling is smiling female.  This can further be combined with disjunction in the row 5 to make either ``non-smiling male'' or ``smiling female''.

\textit{Concept Disjunction:} The last row of \fig{fig:andor} shows EBMs can combine concepts additively (generate images that are concept A or concept B). By constructing sampling using logsumexp, EBMs can sample an image that is ``not smiling male'' or ``smiling female'', where both ``not smiling male'' and ``smiling female'' are specified through the conjunction of energy models of the two concepts.

\textit{Multiple object combination:} We show that our composition operations not only combine object concepts or attributes, but also on the object level. To verify this, we constructed a dataset with one green cube and a large amount background clutter objects (which are not green) in the scene. We train a conditional EBM (conditioned on position) on the dataset. \fig{fig:continual_obj_com1} ``cube 1'' and ``cube 2'' are the generated images conditioned on different positions.
We perform the conjunction operation on the EBMs of ``cube 1'' and ``cube 2'' and use the combined energy model to generate images (row 3). We find that adding two conditional EBMs allows us to selectively generate two different cubes. 
Furthermore, such generation satisfies the constraints of the dataset. For example, when two conditional cubes are too close, the conditionals EBMs are able to default and just generate one cube like the last image in row 3.

\input{figText/continual_obj_comb.tex}

\subsection{Continual Learning}

We evaluate to what extent compositionality in EBMs enables continual learning of new concepts and their combination with previously learned concepts. If we create an EBM for a novel concept, can it be combined with previous EBMs that have never observed this concept in their training data? And can we continually repeat this process? 
\input{figText/continual_eval.tex}
To evaluate this, we use the following methodology on MuJoCo dataset:
1) We first train a position EBM on a dataset of varying positions, but a fixed color and a fixed shape. In experiment, we use shape ``cube'' and color ``purple''. The position EBM allows us generate a purple cube at various positions. (Figure~\ref{fig:continual_obj_com2} row 1).
2) Next we train a shape EBM by training the model in combination with the position EBM to generate images of different shapes at different positions, but without training position EBM. As shown in Figure~\ref{fig:continual_obj_com2} row 2, after combining the position and shape EBMs, the ``sphere'' is placed in the same position as ``cubes'' in row 1 even these ``sphere'' positions never be seen during training.
3) Finally, we train a color EBM in combination with both position and shape EBMs to generate images of different shapes at different positions and colors.  Again we fix both position and shape EBMs, and only train the color model. In Figure~\ref{fig:continual_obj_com2} row 3, the objects with different color have the same position as row 1 and same shape as row 2 which shows the EBM can continually learn different concepts and extrapolate new concepts in combination with previously learned concepts to generate new images.

In \tbl{tbl:continual_tbl}, we quantitatively evaluate the continuous learning ability of our EBM and GAN~\cite{radford2015unsupervised}. Similar to the quantitative evaluation in Section~\ref{sec:composition}, we a train three classifiers for position, shape, color respectively.
For fair comparison, the GAN model is also trained sequentially on the position, shape, and color datasets (with the corresponding position, shape, color and other random attributes set to match the training in EBMs). 

The position accuracy of EBM does not drop significantly when continually learning new concepts (shape and color) which shows our EBM is able to extrapolate earlier learned concepts by combining them with newly learned concepts.  In contrast, while the GAN model is able to learn the attributes of  position, shape and color models given the corresponding dataset. We find the accuracies of position and shape drops significantly after learning color. 
The bad performance shows that GANs cannot combine the newly learned attributes with the previous attributes.

\input{figText/gen_precent.tex}
\input{figText/gen_size_pos.tex}

\subsection{Cross Product Extrapolation}

Humans are endowed with the ability to extrapolate novel concept combinations when only a limited number of combinations were originally observed. For example, despite never having seen a ``purple cube'', a human can compose what it looks like based on the previously observation of ``red cube'' and ``purple sphere''.

 To evaluate the extrapolation ability of EBMs, we construct a dataset of MuJoCo scene images with spheres of all possible sizes appearing only in the top right corner of the scene and spheres of only large size appearing in the remaining positions. The left figure in Figure \ref{fig:gen_precent} shows a qualitative illustration. For the spheres only in the top right corner of the scene, we design different settings. For example, $1\%$ meaning only $1\%$ of positions (starting from the top right corner) that contain all sphere sizes are used for training. At test time, we evaluate the generation of spheres of all sizes at positions that are not seen during the training time. Similar to $1\%$, $10\%$ and $100\%$ mean the spheres of all sizes appears only in the top right $10\%$ and $100\%$ of the scene. The task is to test the quality of generated objects with unseen size and position combinations. This requires the model to extrapolate the learned position and size concepts in novel combinations.

We train two EBMs on this dataset. One is conditioned on the position latent and trained only on large sizes and another is conditioned on the size latent and trained at the aforementioned percentage of positions. Conjunction of the two EBMs is fine-tuned for generation through gradient descent. We compare this composed model with a baseline holistic model conditioned on both position and size jointly. The baseline is trained on the same position and size combinations and optimized directly from the Mean Squared Error between the generated image and real image. Both models use the same architecture and number of parameters are described in the appendix. 

We qualitatively compare the EBM and baseline in \fig{fig:gen_precent}. When sphere of all sizes are only distributed in the $1\%$ of possible locations, both the EBM and baseline have bad performance. This is because the very few combinations of sizes and positions make both models fail in extrapolation. For the $10\%$ setting, our EBM is better than baseline. EBM is able to combine concepts to form images from few combination examples by learning an independent model for each concept factor.  Both EBM and baseline models generate accurate images when given examples of all combinations ($100\%$ setting), but our EBM is closer to ground truth than the baseline.

In \fig{fig:gen_size_pos}, we quantitatively evaluate the extrapolation ability of EBM and the baseline. We train a regression model that outputs both the position and size of a generated sphere image. We compute the error between the predicted size and ground truth size and report it in the first image of \fig{fig:gen_size_pos}. Similarly, we report the position error in the second image. EBMs are able to extrapolate both position and size better than the baseline model with smaller errors. The size errors goes down with more examples of all sphere sizes. For position error, both EBM and the baseline model have smaller errors at $1\%$ data than $5\%$ or $10\%$ data. This result is due to the make-up of the data -- with 1\% data, only 1\% of the rightmost sphere positions have different size annotations, so the models generate large spheres at the conditioned position which are closer to the ground truth position since most positions ($99\%$) are large spheres. 


\subsection{Concept Inference}

Our formulation also allows us to infer concept parameters given a compositional relationship in inputs. For example, given a generated set of of images, each generated by the same underlying concept (conjunction), the likelihood of a concept is given by:
\vspace{-2pt}
\begin{align}
p(x_1, x_2, \ldots, x_n | c) \propto e^{ -\sum_i E(x_i|c) }.
\label{eq:comp_infer}
\end{align}
We can then obtain maximum a posteriori (MAP) estimates of concept parameters by minimizing the logarithm of the above expression. We evaluate inference on an EBM trained on object position, which takes an image and an object position (x,y in 2D) as input and outputs an energy. We analyze the accuracy of such inference in the appendix and find EBMs exhibit both high accuracy and robustness, performing before than a ResNet.

\input{figText/multiview_twocubes.tex}

\myparagraph{Concept Inference from Multiple Observations}
The composition rules in Section~\ref{sec:composition} apply directly to inference.
When given several different views of an object at a particular position with different size, shape, camera view points, and lighting conditions, we can formulate concept inference as inference over a conjunction of multiple positional EBMs.  Each positional EBM takes a different view as input we minimize energy value over positions across the sum of the energies. We use the same metric used above, i.e. Mean Absolute Error, in position inference and find the error in regressing positions goes down when successively giving more images in Figure \ref{fig:multiview}.

\myparagraph{Concept Inference of Unseen Scene with Multiple Objects}
We also investigate the inherent compositionality that emerges from inference on a single EBM generalizing to multiple objects. Given EBMs trained on images of a single object, we test on images with multiple objects (not seen in training). In Figure \ref{fig:twocubes}, we plot the input RGB image and the generated energy maps over all positions in the scene. The ``Two Cubes'' scenes are never seen during training, but the output energy map is still make scene with the bimodality energy distribution. The generated energy map of ``Two Cubes'' is also close to the summation of energy maps of ``Cube 1'' and ``Cube 2'' which shows the EBM is able to infer concepts, such as position, on unseen scene with multiple objects.




%% file: figText/celeba_mujoco_combine.tex
\begin{figure}
     \centering
     \begin{minipage}[b]{0.48\textwidth}
         \centering
         \includegraphics[width=0.9\textwidth]{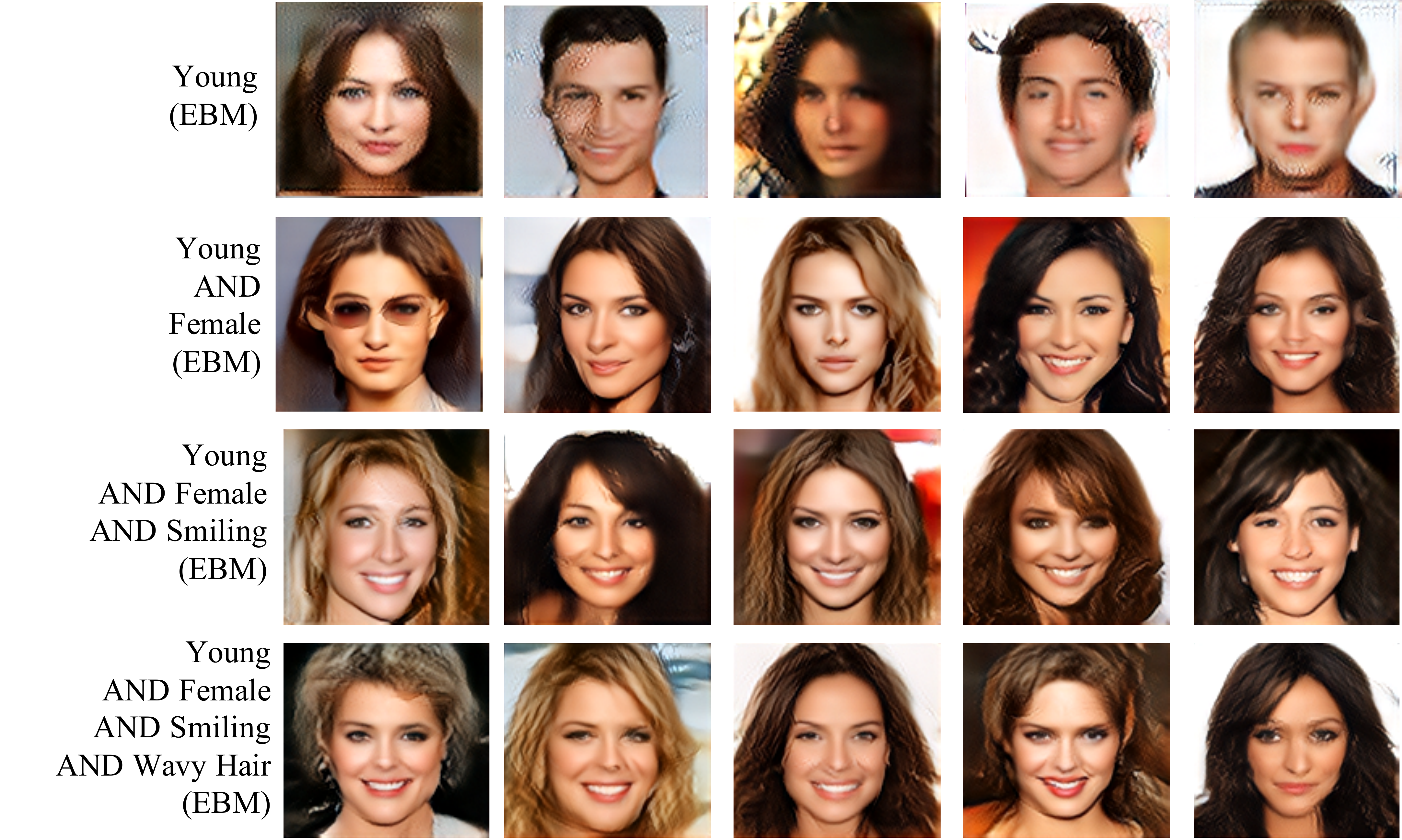}
         \caption{\small Combinations of different attributes on CelebA via concept conjunction. Each row adds an additional energy function. Images on the first row are conditioned on young, while images on the last row are conditioned on young, female, smiling, and wavy hair.}
        \label{fig:celeba_combine}
     \end{minipage}
     \hfill
     \begin{minipage}[b]{0.48\textwidth}
         \centering
         \includegraphics[width=0.9\textwidth]{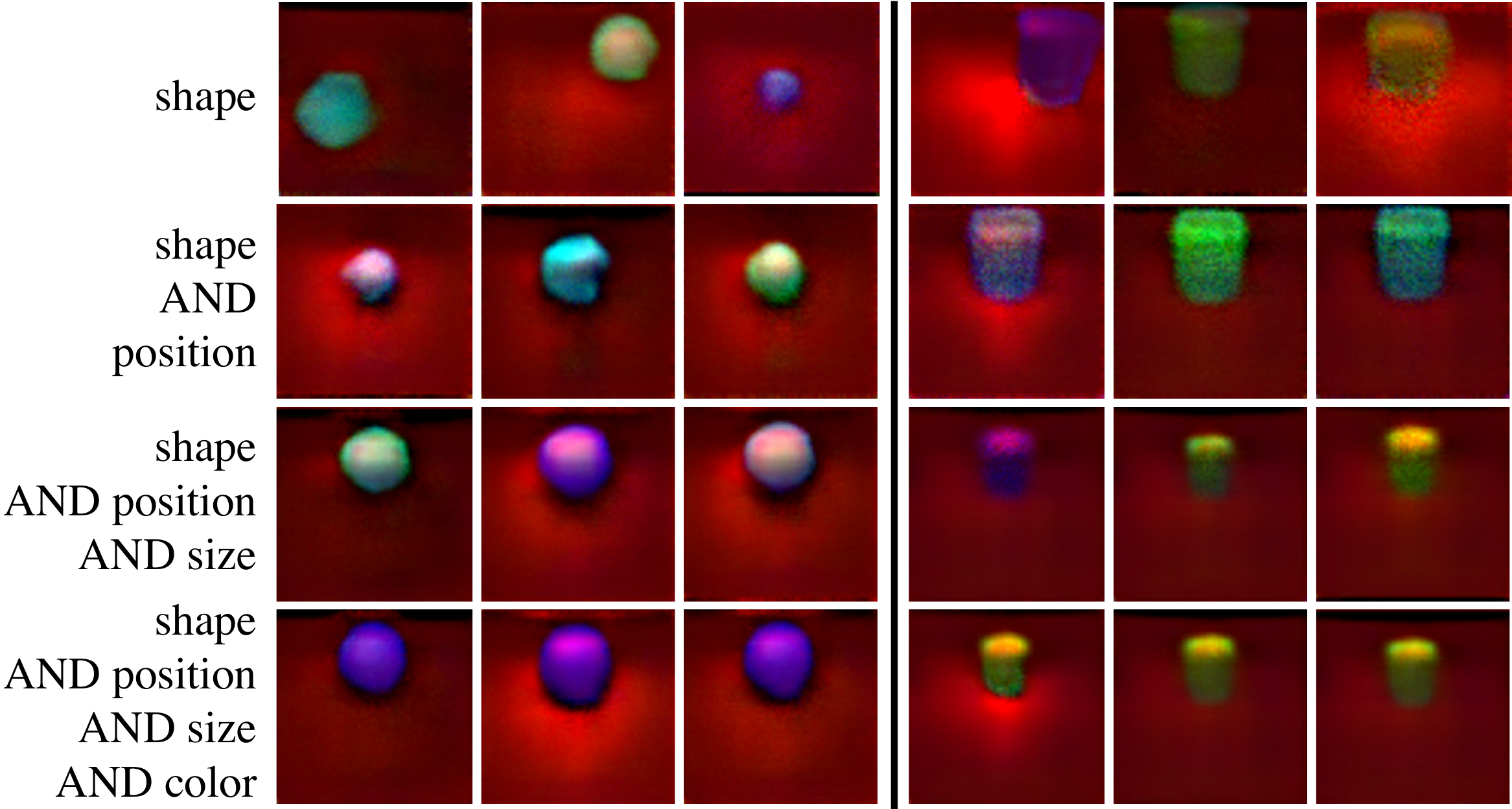}
        \caption{\small Combinations of different attributes on MuJoCo via concept conjunction. Each row adds an additional energy function. Images on the first row are only conditioned on shape, while images on the last row are conditioned on shape, position, size, and color. The left part is the generation of a sphere shape and the right is a cylinder.}
        \label{fig:mujoco_combine}
    \end{minipage}
    \vspace{-10pt}
\end{figure}

%% file: figText/disjunction_conjunction.tex

\begin{figure}
    \begin{minipage}{0.4\textwidth}
    \begin{center}
    \includegraphics[width=\textwidth]{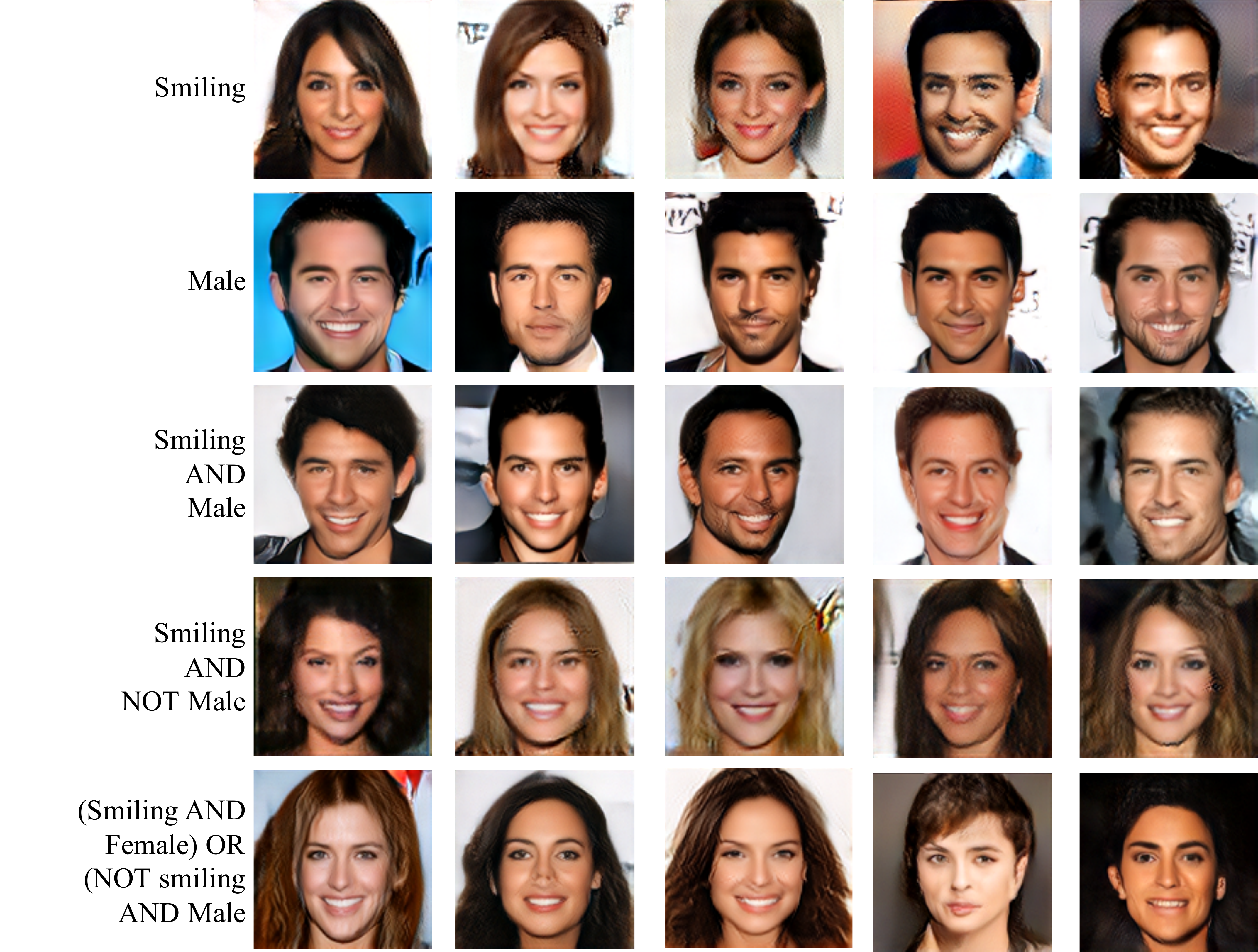}
    \end{center}
    \vspace{-10pt}
    \caption{\small Examples of recursive compositions of disjunction, conjunction, and negation on the CelebA dataset. 
    }
    \label{fig:andor}
    \vspace{-10pt}
    \end{minipage}\hfill
    \begin{minipage}{0.55\textwidth}
    \begin{subtable}{\textwidth}
    \begin{center}
    \resizebox{\textwidth}{!}{
    \begin{tabular}{l|cc}
    \toprule
    \bf Model & \bf Pos Acc & \bf Color Acc  \\ 
    \midrule
    Color & 0.128 & 0.997 \\  
    Pos & 0.984 & 0.201 \\    
    Pos $\&$ Color & 0.801 &  0.8125\\    
    Pos $\&$ ($\neg$ Color) & 0.872 & 0.096\\  
  ($\neg$ Pos) $\&$ Color & 0.033 & 0.971 \\  
    Color \citep{vedantam2017generative} & 0.132 & 0.333 \\  
    Pos \citep{vedantam2017generative} & 0.146 & 0.202 \\    
    Pos $\&$ Color \citep{vedantam2017generative} & 0.151 &  0.342\\    
    \midrule
    \bf Model & \bf Pos 1 Acc & \bf Position 2 Acc\\ 
    \midrule
    Pos 1 & 0.875  &  0.0\\
    Pos 2 & 0.0 &  0.817\\
    Pos 1 $|$ Pos 2 & 0.432 &  0.413 \\
    \midrule
    \bf Model & \bf Pos/Color 1  Acc & \bf Pos 2/Color 2 Acc\\ 
    \midrule
    Pos 1 $\&$ Color 1 & 0.460  &  0.0\\
    Pos 2 $\&$ Color 2 & 0.0 &  0.577\\
    (Pos 1 $\&$ Color 1) $|$ (Pos 2 $\&$ Color 2) & 0.210  &  0.217 \\
    \bottomrule
    \end{tabular}%
    }
    \end{center}
    \end{subtable}
    \captionof{table}{\small Quantitative evaluation of conjunction ($\&$), disjunction ($|$) and negation ($\neg$) generations on the Mujoco Scenes dataset using an EBM or  the approach in \citep{vedantam2017generative}. Position = Pos. Each individual attribute (Color or Position ) generation is a individual EBM. (Acc: accuracy) Standard error is close to 0.01 for all models. }
    \label{tbl:disjunction_tbl}
    \end{minipage}
\end{figure}

%% file: figText/continual_obj_comb.tex
\begin{figure*}
\centering
\begin{minipage}[t]{0.47\textwidth}
    \includegraphics[height=3.2cm, width=6.1cm]{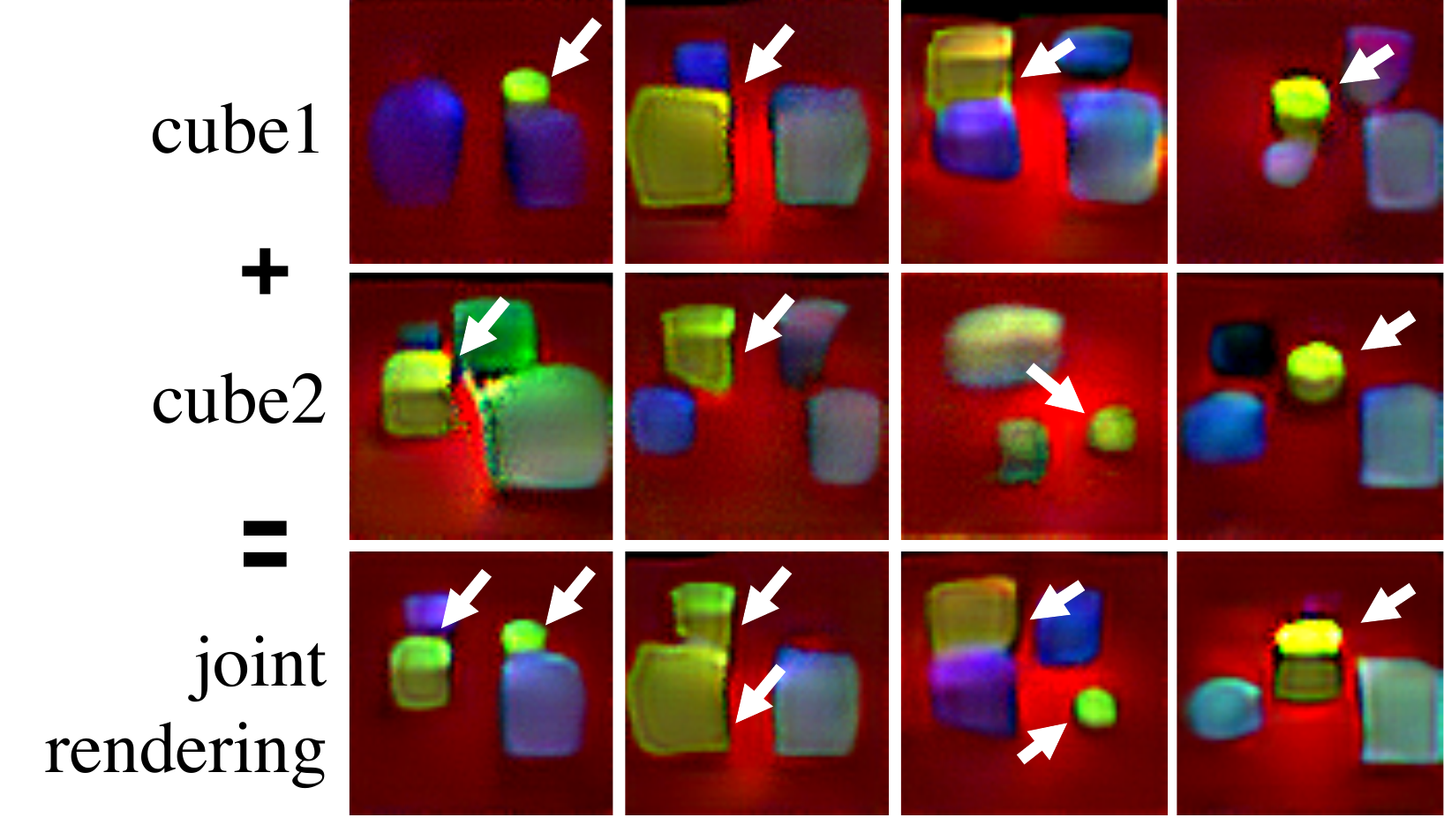}
    \vspace{-5pt}
    \caption{\small Multi-object compositionality with EBMs. An EBM is trained to generate a green cube at location in a scene alongside other objects. At test time, we sample from the conjunction of two EBMs conditioned on different positions and sizes (cube 1 and 2) and generates cubes at both locations. Two cubes are merged into one if they are too close (last column).}
    \label{fig:continual_obj_com1}
\end{minipage}\hfill
\begin{minipage}[t]{0.49\textwidth}
    \includegraphics[height=3.2cm, width=6.4cm]{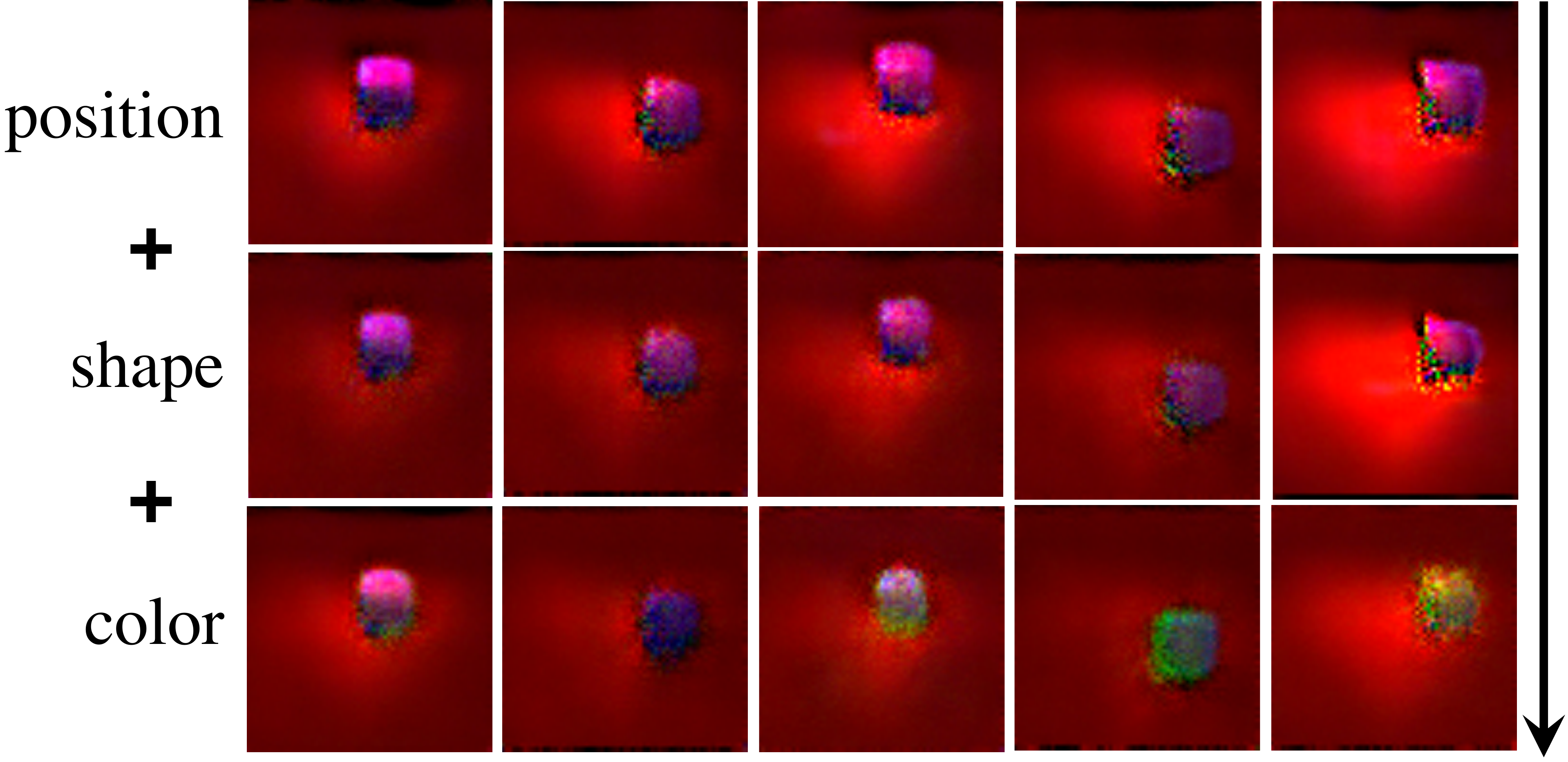}
    \vspace{-5pt}
    \caption{\small Continual learning of concepts. A position EBM is first trained on one shape (cube) of one color (purple) at different positions (first row). A shape EBM is then trained on different shapes of one fixed color (purple) (second row). Finally, a color EBM is trained on shapes of many colors (third row). EBMs learn to combine concepts to many shapes (cube, sphere), colors and positions.}
    \label{fig:continual_obj_com2}
\end{minipage}\hfill
\vspace{-10pt}
\end{figure*}

%% file: figText/continual_eval.tex
\begin{wraptable}{r}{.6\linewidth}
\vspace{-10pt}
\small\centering\vspace{-2pt}
\caption{\small Quantitative evaluation of continual learning. A position EBM is first trained on ``purple'' ``cubes'' at different positions. A shape EBM is then trained on different ``purple'' shapes. Finally, a color EBM is trained on shapes of many colors with Earlier EBMs are fixed and combined with new EBMs. We compare with a GAN model~\cite{radford2015unsupervised} which is also trained on the same position, shape and color dataset. EBMs is better at continually learning new concepts and remember the old concepts. (Acc: accuracy) 
}
\label{tbl:continual_tbl}
\vspace{-5pt}
\resizebox{0.6\textwidth}{!}{
\begin{tabular}{l|ccc}
\toprule
\bf Model & \bf Position Acc & \bf Shape Acc  & \bf Color Acc \\ 
\midrule
EBM (Position) & 0.901 & - & -\\ 
EBM (Position + Shape) & 0.813 & 0.743 & - \\  
EBM (Position + Shape + Color) & 0.781 & 0.703 & 0.521 \\ 
\midrule
GAN (Position) & 0.941 & - & - \\
GAN (Position + Shape) & 0.111 & 0.977 & - \\
GAN (Position + Shape + Color) & 0.117 & 0.476 & 0.984\\
\bottomrule
\end{tabular}
}
\vspace{-10pt}
\end{wraptable}


%% file: figText/gen_precent.tex
\begin{figure*}[h]
\begin{center}

\begin{minipage}{.2\textwidth}
    \centering
    \includegraphics[height=0.9\textwidth]{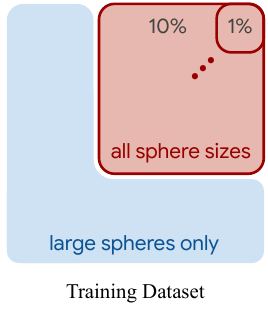}
    \vspace{4pt} 
\end{minipage}
\begin{minipage}{.7\textwidth}
    \centering
    \includegraphics[width=1\textwidth]{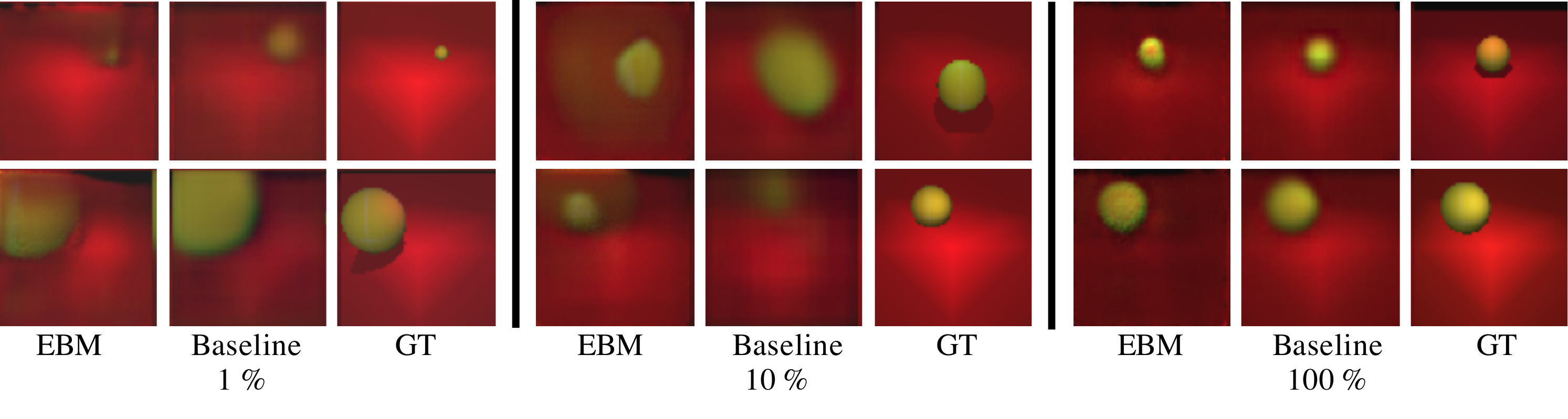}
\end{minipage}

\end{center}
\vspace{-10pt}
\caption{\small Cross product extrapolation. Left: the spheres of all sizes only appear in the top right corner (1\%, 10\%, \dots) of the scene and the remaining positions only have large size spheres. Right: generated images of novel size and position combinations using EBM and the baseline model.}
\label{fig:gen_precent}
\end{figure*}

%% file: figText/gen_size_pos.tex
\begin{figure}[t]
\begin{center}
\includegraphics[width=0.48\textwidth]{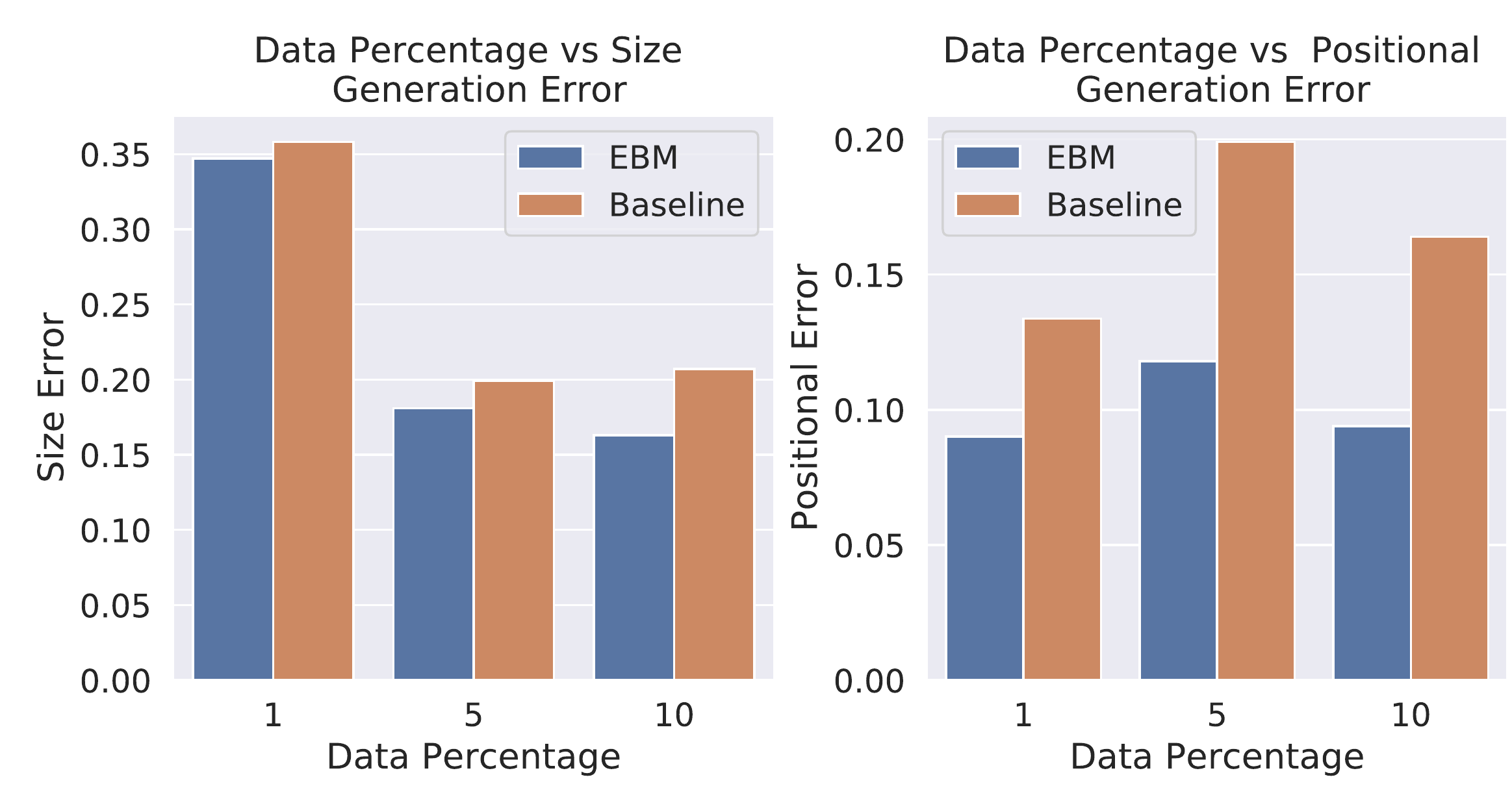}
\end{center}
\caption{\small Cross product extrapolation results with respect to the percentages of areas on the top right corner.
EBM has lower size and position errors which means EBM is able to extrapolate better with less data than the baseline model.}
\label{fig:gen_size_pos}

\end{figure}

%% file: figText/multiview_twocubes.tex
\begin{figure*}
    \begin{minipage}[t]{0.47\textwidth}
    \begin{center}
    \includegraphics[width=0.78\textwidth,height=0.5\textwidth]{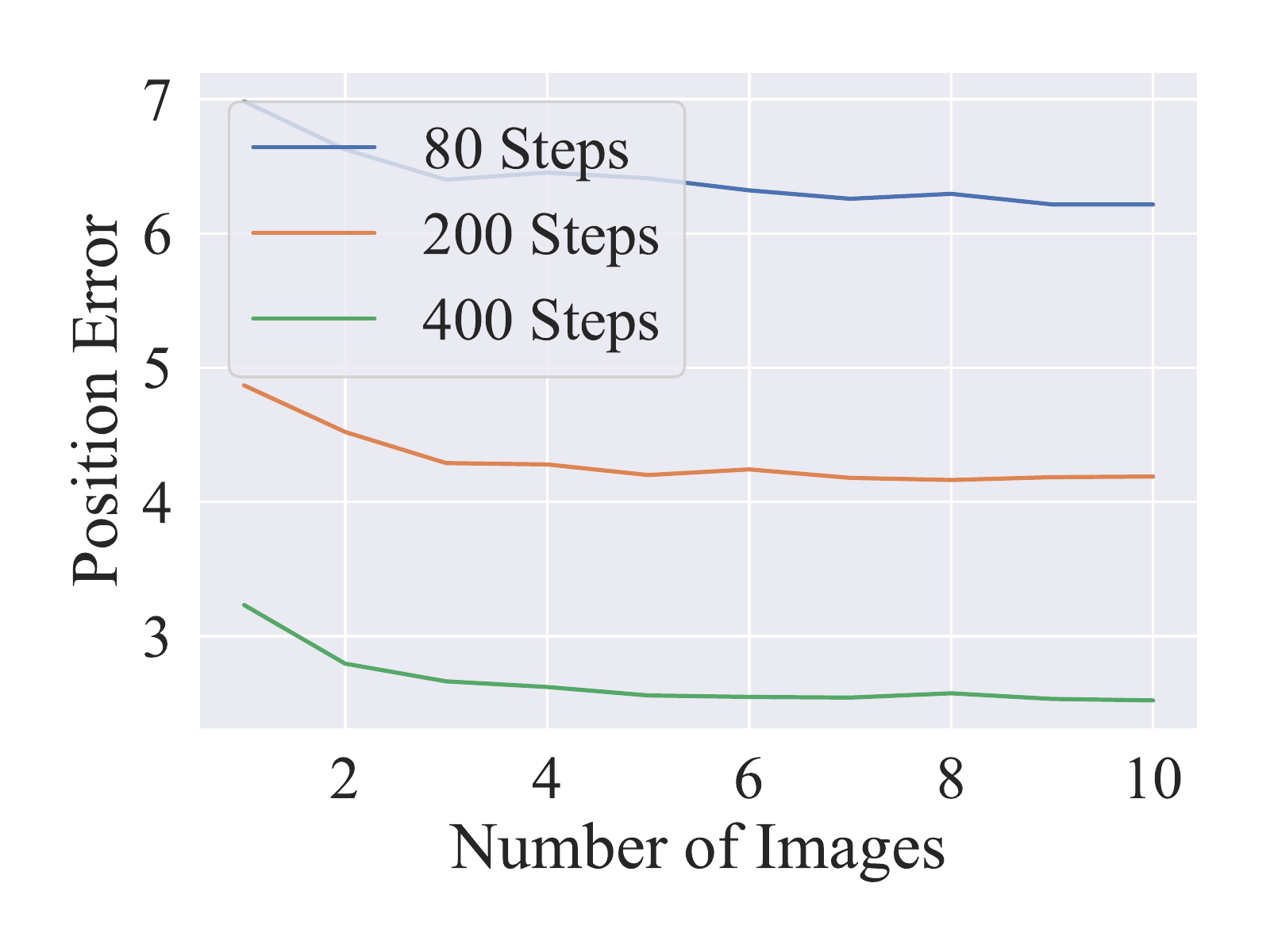}
    \end{center}
    \vspace{-10pt}
    \caption{\small Concept inference from multiple observations. Multiple images are generated under different size, shape, camera view points, and lighting conditions.
    The position prediction error decreases when the number of input images increases with different Langevin Dynamics sampling steps for training.}
    \label{fig:multiview}
    \end{minipage}
    \hfill
    \begin{minipage}[t]{0.48\textwidth}
    \begin{center}
    \includegraphics[width=0.67\linewidth]{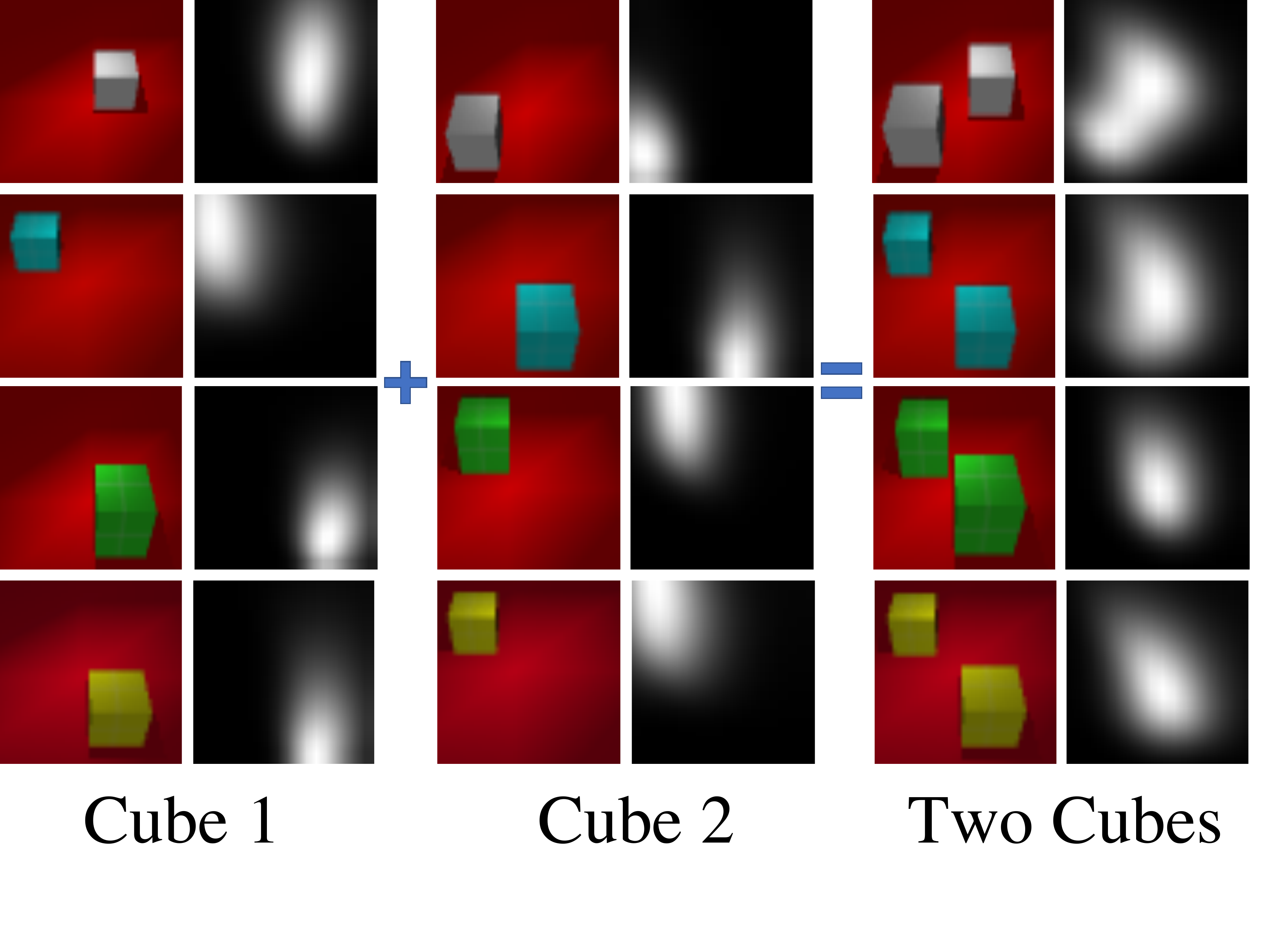}
    \end{center}
    \vspace{-10pt}
    \caption{\small Concept inference of multiple objects with EBM trained on single cubes and tested on two cubes. The color images are the input and the gray images are the output energy map over all positions. The energy map of two cubes correctly shows the bimodality which is close to the summation of the front two energy maps.}
     \label{fig:twocubes}
    \end{minipage}
    \vspace{-20pt}
\end{figure*}

%% file: text/discussion.tex
\vspace{-2pt}
\section{Conclusion}
\vspace{-2pt}
In this paper, we demonstrate the potential of EBMs for both compositional generation and inference. We show that EBMs support composition on both the factor and object level, unifying different perspectives of compositionality and can recursively combine with each other. We further showcase how this composition can be applied to both continually learn and compositionally infer underlying concepts. We hope our results inspire future work in this direction.




%% file: text/acknowledgement.tex
\section{Acknowledgement}
We should like to thank Jiayuan Mao for reading and providing feedback on the paper and both Josh Tenenbaum and Jiayuan Mao for helpful feedback on the paper.

%% file: text/broader_impacts.tex
\section{Broader Impacts}
We believe that compositionality is a crucial component of next generation AI systems. Compositionality enables system to synthesize and combine 
knowledge from different domains to tackle the problem in hand. Our proposed method is step towards more composable deep learning models. A truly compositional system has many positive societal benefits, potentially enabling a intelligent and flexible robots that can selectively recruit different skills learned for the task on hand, or super-human synthesis of scientific knowledge that can further progress of scientific discovery. At the same time, there remain unanswered ethical problems about any such next generation AI system.



%% file: text/appendix.tex
\newpage
\section{Appendix}
\label{appendix}

\subsection{Inference}

To evaluate the inference ability of EBMs, we generate a new MuJoCo Scene dataset for training and testing. Each scene has varying lighting conditions with one object, either sphere or cube, at all possible positions and some sizes.  We build several different test datasets to evaluate generalization if models. The easiest one is \emph{``Test''} which has the same data distribution with the training dataset. The \emph{``Size''} test dataset contains objects twice the size of training objects. \emph{``Color''} dataset has object colors never been seen during training. \emph{``Light''} is a test dataset with different light sources and \emph{``Type''} dataset consists of cylinder images while the training images are only spheres or cubes.

We evaluate inference on an EBM trained on object position, which takes an image and an object position (x,y in 2D) as input and outputs an energy. We iterate densely over all positions (20 by 20 grid of positions) and select the position with the minimal energy as our inference result. We evaluate this result by computing the Mean Absolute Error between the predicted position and ground truth object position.

We compare EBMs with two baseline models, ResNet model \citep{he2016deep} (with the same architecture as EBM) and PixelCNN \citep{oord2016pixel}.
\tbl{tbl:compositional_inference} shows the comparison results using different number of Langevin Dynamics sampling steps ($k$ in Equation 3 in the main text). We find that inference in EBMs is able to generalize well to different out of distribution datasets such as Color, Light, Size and Type. A large number of Langevin sampling steps also improves performance, with a large number of steps of training exhibiting both better training accuracy and generalization performance.

\input{figText/diff_cond.tex}

\subsection{Partition Function}
\label{sect:partition}
\begin{figure}[H]
\begin{center}
\centering
\includegraphics[width=1\textwidth]{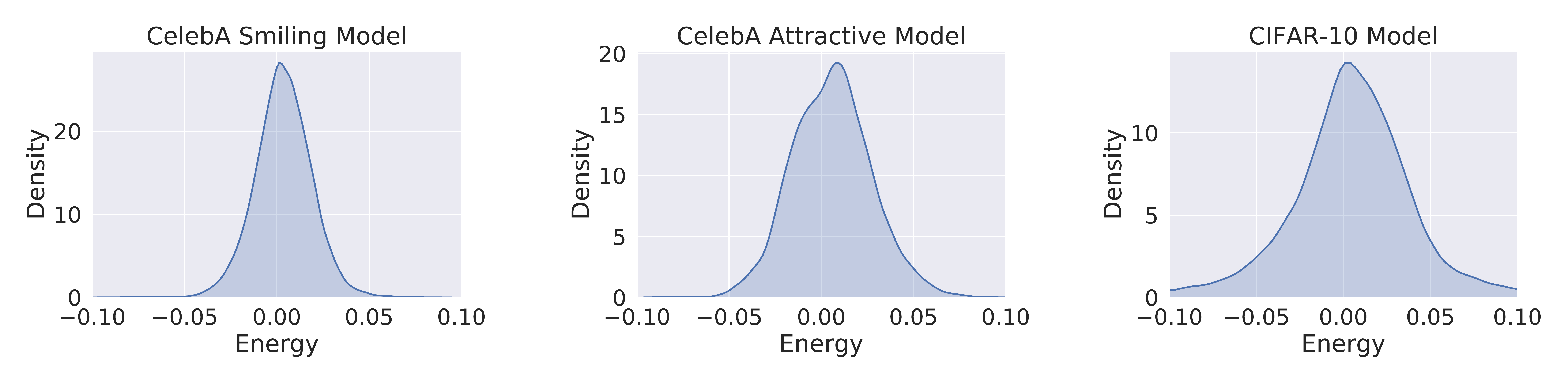}
\end{center}
\vspace{-10pt}
\caption{\small Energy histograms of models trained on CelebA smiling (left), CelebA attractive (middle) and pretrained CIFAR-10 model from \citep{du2019implicit} (right). Each EBM we evaluate have different architectures but still have similar histograms.}
\label{fig:partition_evaluation}
\vspace{-10pt}
\end{figure}

We estimate the magnitude of the partition function of an EBM by evaluating the energy it assigns to all data points it is trained on, and plot the resultant histogram of energies. \fig{fig:partition_evaluation} shows that the EBMs we train have similar histograms due to a combination of  L2 normalization and spectral normalization. The EBMs we evaluated have different architectures but similar histograms.

Specifically, in \fig{fig:partition_evaluation}, we compare the energy histogram of a CelebA model trained on either smiling or attractive histograms as well as the CIFAR-10 model from \citep{du2019implicit}.  We find that all energy histograms are similar, exhibiting minimum and maximum energies between -0.01 and 0.01. This is true even for the  CIFAR-10 model  which uses a significantly different dataset and architecture. 

\subsection{Analysis of Mismatch of Partition Function on Disjunction}

\begin{wrapfigure}{R}{0.3\textwidth}
\vspace{-10pt}
\begin{center}
\includegraphics[width=0.9\linewidth]{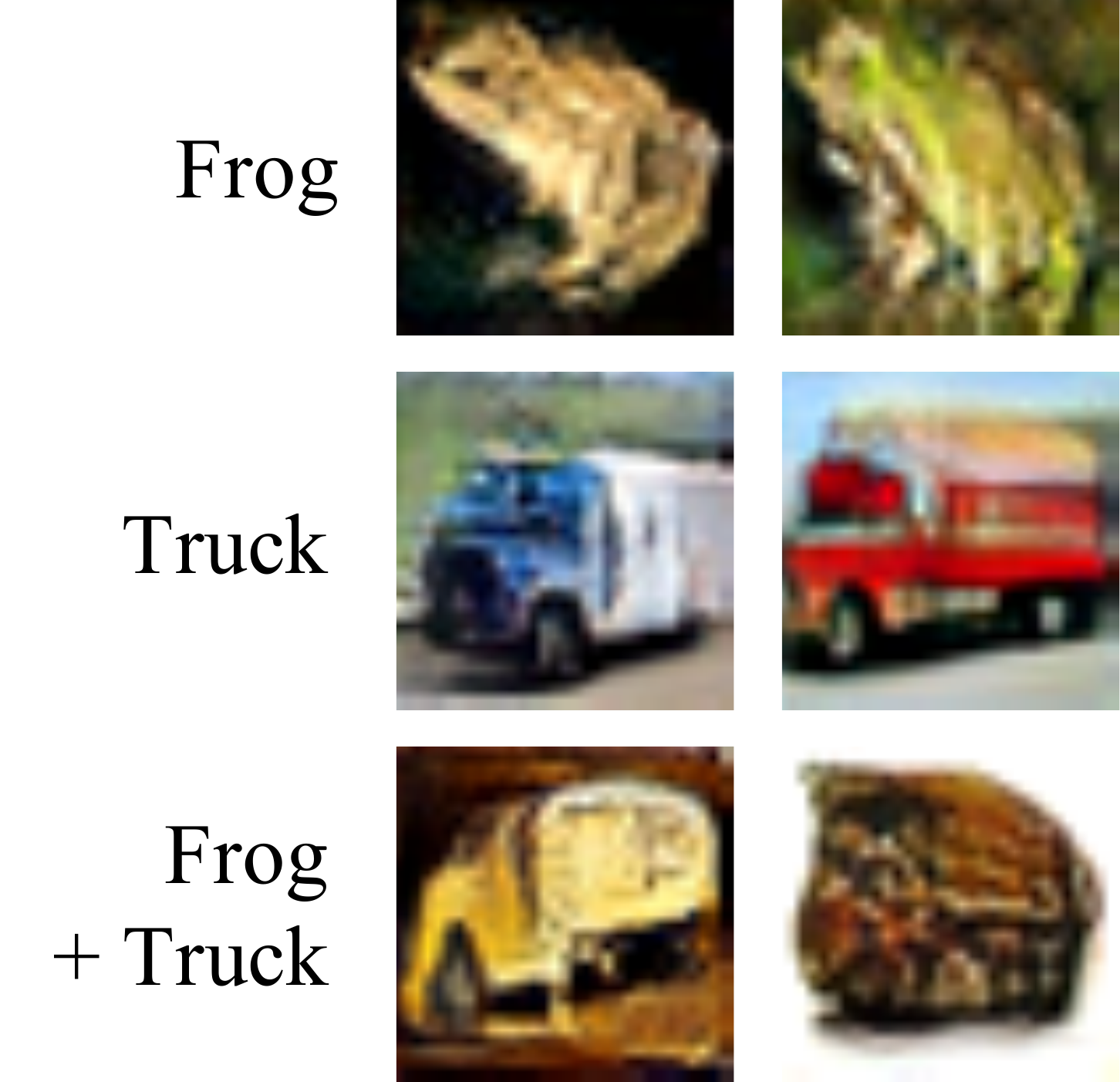}
\end{center}
\caption{\small Hybrid combinations of frog and truck EBMs.
}
\label{fig:hybrid}
\vspace{-20pt}
\end{wrapfigure}
In scenarios where partition functions are different, our defined disjunction operator does not fail drastically. If two un-normalized probability distributions have partition function values of $w_1$ and $w_2$ then models will be sampled with proportion $\tfrac{w_1}{w_1+w_2}$ and $\tfrac{w_2}{w_1+w_2}$, which is not a dramatic failure in disjunction.


\subsection{Disjoint Compositionality Results}
We further evaluate compositionality when conditioned factors are mutually disjoint from each other. In particular, we train EBM models on frog and truck image classes in CIFAR-10. In \fig{fig:hybrid}, we illustrate resulting generations. We find that when conditioning on both classes, our resultant generations exhibit characteristics of each individual class.

\subsection{Discussion on Other Generative Models}
To sample from the conjunction/disjunction/negation of seperate probability distributions, MCMC must be run.  Other generative models, such as autoregressive models, can also support MCMC, but we find that in practice other generative models do not sample well under gradient based MCMC. 

\begin{figure}[H]
\begin{subfigure}[t]{0.45\textwidth}
\centering
\includegraphics[width=0.8\textwidth]{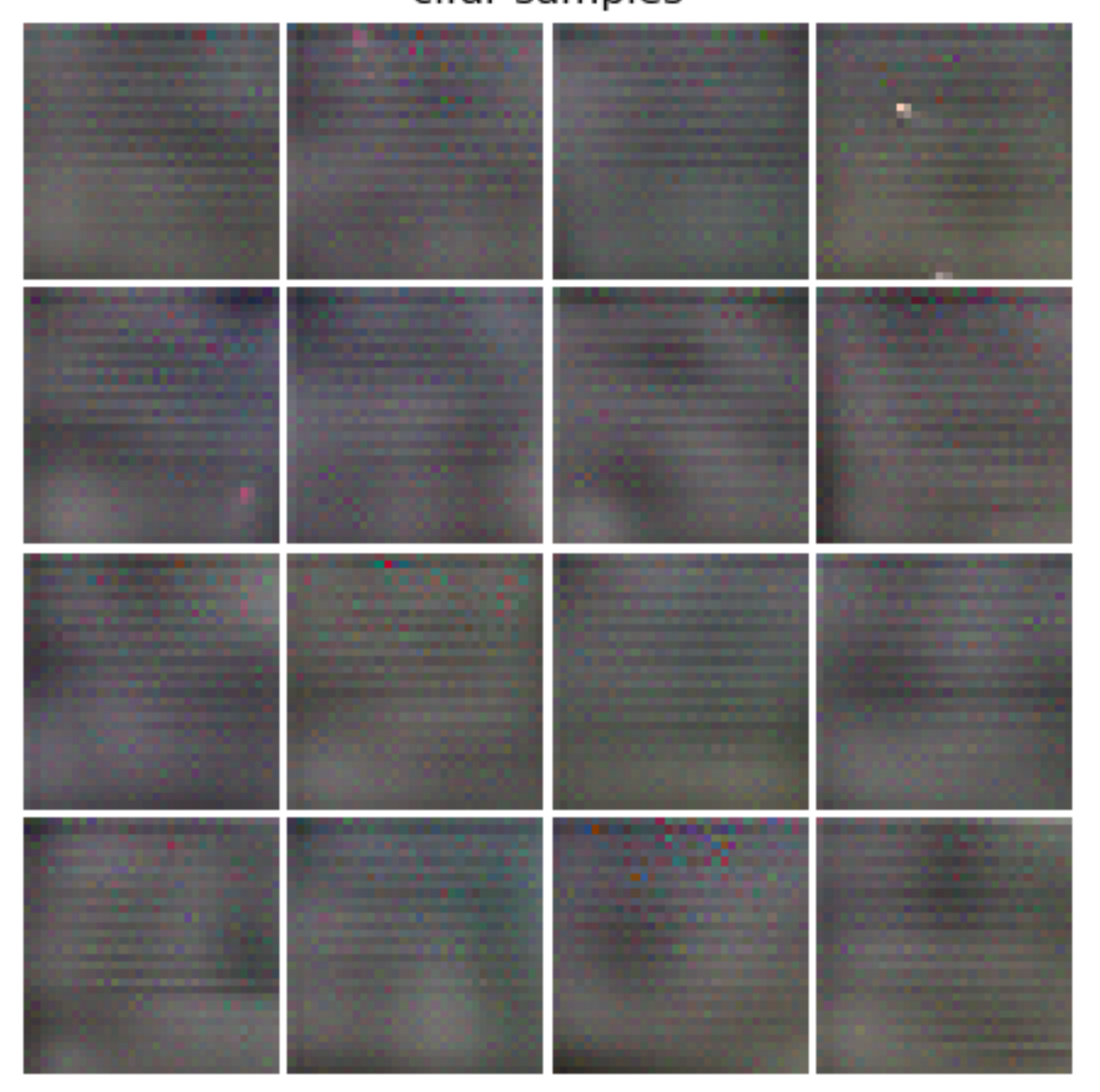}
\caption{Samples Generated from Langevin Sampling on PixelCNN++ model from \citep{salimans2017pixelcnn++}. }
\label{fig:pixelcnn_langevin}
\end{subfigure}
\begin{subfigure}[t]{0.45\textwidth}
\centering
\includegraphics[width=0.8\textwidth]{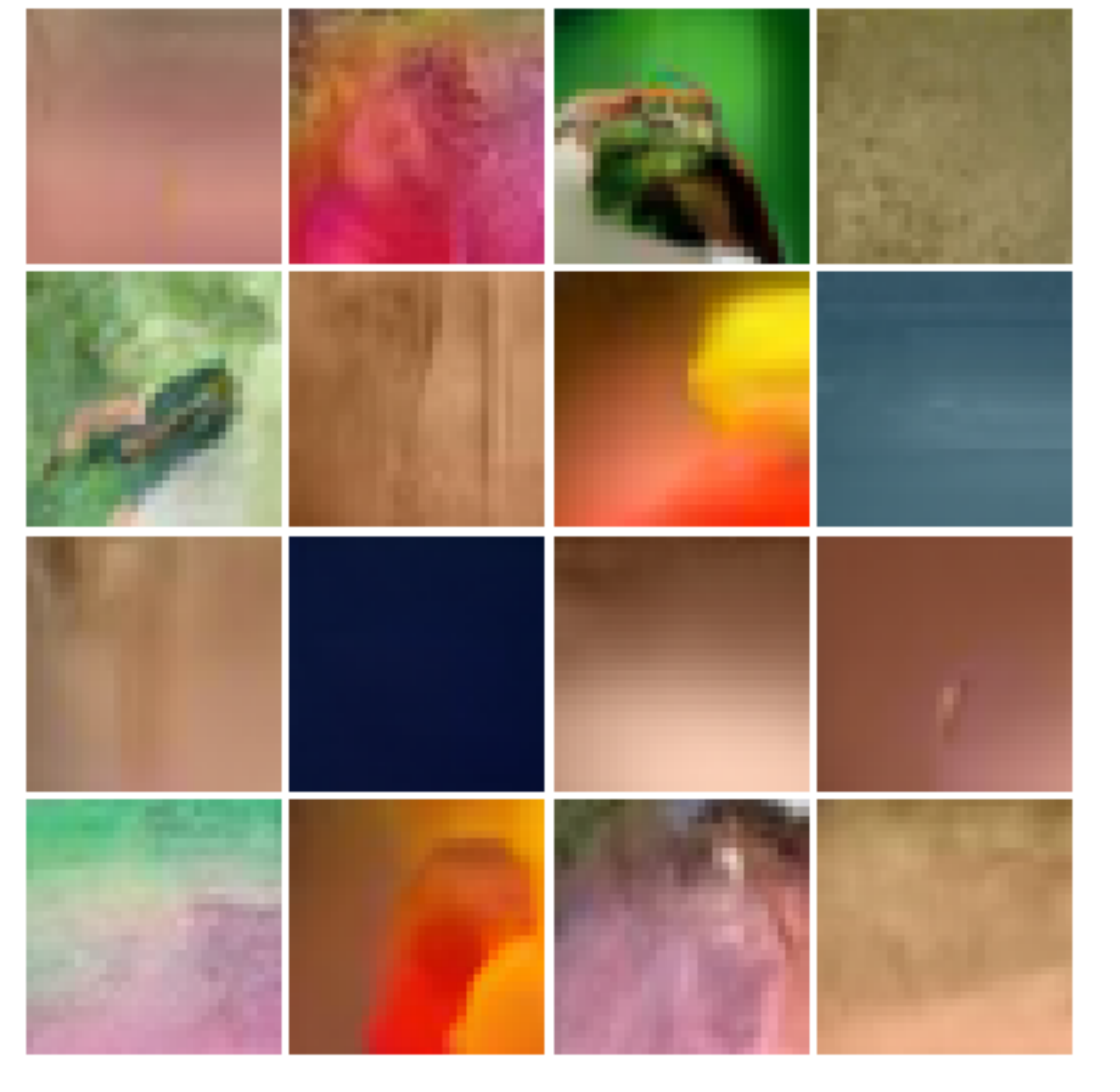}
\caption{Samples Generated from Autoregressive Sampling on PixelCNN++ model from \citep{salimans2017pixelcnn++}.}
\label{fig:pixelcnn_auto}
\end{subfigure}%
\caption{Comparison on samples generated from different sampling scenes on PixelCNN++ model from \citep{salimans2017pixelcnn++}. We note that Langevin sampling, while not making realistic samples, generate \textbf{higher} likelihood samples than those from autoregressive sampling.}
\label{fig:pixelcnn_sample}
\end{figure}

We considered Langevin based sampling on the pretrained CIFAR-10 unconditional PixelCNN++ model \citep{salimans2017pixelcnn++} in \fig{fig:pixelcnn_sample}. While both sampling schemes generate images with similar likelihoods (with Langevin sampling creating higher likelihood samples), we find images generated from Langevin sampling are significantly worse than those generated from autoregressive sampling. We speculate that EBMs fit the MCMC sampling procedure better than other models since EBMs are trained with MCMC inference, and are thus less susceptible to adversarial modes.


\subsection{Models}
\label{appendix:model}

\begin{figure}[H]
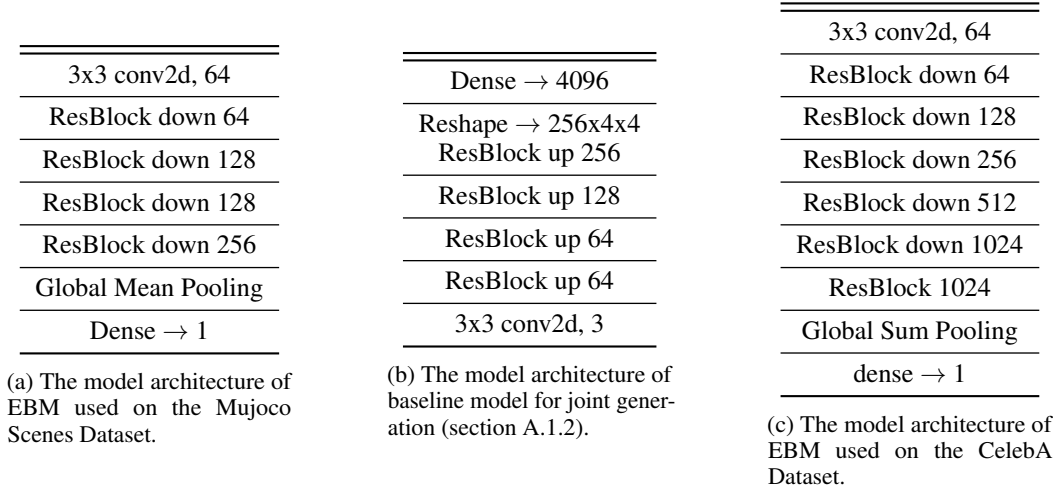

\begin{subfigure}[t]{0.27\textwidth}
\centering
\begin{tabular}{c}
    \toprule
    \toprule
    3x3 conv2d, 64 \\
    \midrule
    ResBlock down 64 \\
    \midrule
    ResBlock down 128 \\
    \midrule
    ResBlock down 128 \\
    \midrule
    ResBlock down 256 \\
    \midrule
    Global Mean Pooling \\ 
    \midrule
    Dense $\rightarrow$ 1 \\ 
    \bottomrule
\end{tabular}
\caption{The model architecture of EBM used on the Mujoco Scenes Dataset.}
\label{fig:mujocomodel}
\end{subfigure}
\hfill
\begin{subfigure}[t]{0.27\textwidth}
\centering
\begin{tabular}{c}
    \toprule
    \toprule
    Dense $\rightarrow$ 4096\\
    \midrule
    Reshape $\rightarrow$ 256x4x4 \\
    ResBlock up 256 \\
    \midrule
    ResBlock up 128 \\
    \midrule
    ResBlock up 64 \\
    \midrule
    ResBlock up 64 \\
    \midrule
    3x3 conv2d, 3\\ 
    \bottomrule
\end{tabular}
\caption{The model architecture of baseline model for joint generation (section A.1.2).}
\label{fig:baseline3_4}
\end{subfigure}
\hfill
\begin{subfigure}[t]{0.27\textwidth}
\centering
\begin{tabular}{c}
    \toprule
    \toprule
    3x3 conv2d, 64 \\
    \midrule
    ResBlock down 64 \\
    \midrule
    ResBlock down 128 \\
    \midrule
    ResBlock down 256 \\
    \midrule
    ResBlock down 512 \\ 
    \midrule
    ResBlock down 1024 \\
    \midrule
    ResBlock 1024 \\
    \midrule
    Global Sum Pooling \\ 
    \midrule
    dense $\rightarrow$ 1\\
    \bottomrule
\end{tabular}
\caption{The model architecture of EBM used on the CelebA Dataset.}
\label{fig:celebamodel}
\end{subfigure}
\label{fig:architecture}
\caption{Architecture of models on different datasets.}
\end{figure}

We detail the EBM architectures used for the Mujoco Scenes images in \fig{fig:mujocomodel} and for the Celeba 128x128 images in \fig{fig:celebamodel}. The baseline model used for comparisons in section 3.4 is in \fig{fig:baseline3_4}.

\subsection{Training Details/Hyperparameters/Source Code}
\label{appendix:training_detail}
 Models trained on Mujoco Scenes and CelebA datasets use the Adam optimizer with the learning rate 3e-4, first order moment 0.0, and second order moment 0.999. The batch size is 128. The replay buffer size is 50000 with a 5\% replacement rate. Spectral normalization is applied to models with a step size of 100 for each Langevin dynamics step. We use 60 steps of Langevin sampling per training iteration for the CelebA dataset and 80 steps of Langevin sampling per training iteration for the Mujoco Scenes dataset. We use the Swish activation to train our models (as noted in \citep{du2019implicit}), and find that it greatly stabilizes and speeds up training of models.

%% file: figText/diff_cond.tex
\begin{table}[h]
\caption{\small Position error on different test datasets. ``Test'' has the same data distribution with training set. Other datasets change one environmental parameter, e.g. color, size, type, and light, which are unseen in the training set. ``Avg'' is the average error of ``Color'', ``Light'', ``Size'', and ``Type''. ``Steps''indicates the number of sampling steps used to train the EBMs. EBMs are able to generalize better on unseen datasets. Larger number of sampling steps significantly decrease overall EBM error.}
\label{tbl:compositional_inference}
\begin{center}
\begin{tabular}{ccccccc|c}
\toprule
\bf Model & \bf Steps & \bf Color  & \bf Light & \bf Size & \bf Type & \bf Avg & \bf Test \\ 
\midrule
EBM & 200 & 10.899 & 6.307 & 8.431 & 6.304 & 7.985 & 3.903 \\  
EBM & 400 & \textbf{4.084} & \textbf{4.033} & \textbf{6.853} & \textbf{3.694} & \textbf{4.666} & \textbf{2.917} \\ 
Resnet & - & 20.002 & 5.881 & 10.378 & 6.310 & 10.643 & 3.635 \\
PixelCNN & - & 60.607 & 58.589 & 33.889 & 48.138 & 50.306 & 43.460 \\
\bottomrule
\end{tabular} %
\end{center}
\vspace{-10pt}
\end{table}

%% file: icml2020_conference.bbl
\begin{thebibliography}{38}
\providecommand{\natexlab}[1]{#1}
\providecommand{\url}[1]{\texttt{#1}}
\expandafter\ifx\csname urlstyle\endcsname\relax
  \providecommand{\doi}[1]{doi: #1}\else
  \providecommand{\doi}{doi: \begingroup \urlstyle{rm}\Url}\fi

\bibitem[Andreas(2019)]{andreas2019measuring}
Jacob Andreas.
\newblock Measuring compositionality in representation learning.
\newblock \emph{arXiv preprint arXiv:1902.07181}, 2019.

\bibitem[Benaim et~al.(2019)Benaim, Khaitov, Galanti, and
  Wolf]{Benaim2019DomainIntersectionDifference}
Sagie Benaim, Michael Khaitov, Tomer Galanti, and Lior Wolf.
\newblock Domain intersection and domain difference.
\newblock In \emph{ICCV}, 2019.

\bibitem[Du and Mordatch(2019)]{du2019implicit}
Yilun Du and Igor Mordatch.
\newblock Implicit generation and generalization in energy-based models.
\newblock \emph{arXiv preprint arXiv:1903.08689}, 2019.

\bibitem[Du et~al.(2019)Du, Lin, and Mordatch]{du2019plan}
Yilun Du, Toru Lin, and Igor Mordatch.
\newblock Model based planning with energy based models.
\newblock \emph{CoRL}, 2019.

\bibitem[Du et~al.(2020)Du, Li, Tenenbaum, and Mordatch]{du2020cd}
Yilun Du, Shuang Li, Joshua Tenenbaum, and Igor Mordatch.
\newblock Improved contrastive divergence training of energy based models.
\newblock \emph{arXiv preprint arXiv:2012.01316}, 2020.

\bibitem[Eslami et~al.(2018)Eslami, Rezende, Besse, Viola, Morcos, Garnelo,
  Ruderman, Rusu, Danihelka, Gregor, et~al.]{eslami2018neural}
SM~Ali Eslami, Danilo~Jimenez Rezende, Frederic Besse, Fabio Viola, Ari~S
  Morcos, Marta Garnelo, Avraham Ruderman, Andrei~A Rusu, Ivo Danihelka, Karol
  Gregor, et~al.
\newblock Neural scene representation and rendering.
\newblock \emph{Science}, 360\penalty0 (6394):\penalty0 1204--1210, 2018.

\bibitem[Fodor and Lepore(2002)]{fodor2002compositionality}
Jerry~A Fodor and Ernest Lepore.
\newblock \emph{The compositionality papers}.
\newblock Oxford University Press, 2002.

\bibitem[Greff et~al.(2019)Greff, Kaufmann, Kabra, Watters, Burgess, Zoran,
  Matthey, Botvinick, and Lerchner]{greff2019multi}
Klaus Greff, Rapha{\"e}l~Lopez Kaufmann, Rishab Kabra, Nick Watters, Chris
  Burgess, Daniel Zoran, Loic Matthey, Matthew Botvinick, and Alexander
  Lerchner.
\newblock Multi-object representation learning with iterative variational
  inference.
\newblock \emph{arXiv preprint arXiv:1903.00450}, 2019.

\bibitem[Gregor et~al.(2015)Gregor, Danihelka, Graves, Rezende, and
  Wierstra]{Gregor2015DRAW}
Karol Gregor, Ivo Danihelka, Alex Graves, Danilo~Jimenez Rezende, and Daan
  Wierstra.
\newblock Draw: A recurrent neural network for image generation.
\newblock \emph{arXiv preprint arXiv:1502.04623}, 2015.

\bibitem[He et~al.(2016)He, Zhang, Ren, and Sun]{he2016deep}
Kaiming He, Xiangyu Zhang, Shaoqing Ren, and Jian Sun.
\newblock Deep residual learning for image recognition.
\newblock In \emph{Proceedings of the IEEE conference on computer vision and
  pattern recognition}, pages 770--778, 2016.

\bibitem[Higgins et~al.(2017)Higgins, Matthey, Pal, Burgess, Glorot, Botvinick,
  Mohamed, and Lerchner]{Higgins2017Beta}
Irina Higgins, Loic Matthey, Arka Pal, Christopher~P Burgess, Xavier Glorot,
  Matthew Botvinick, Shakir Mohamed, and Alexander Lerchner.
\newblock Beta-vae: Learning basic visual concepts with a constrained
  variational framework.
\newblock In \emph{ICLR}, 2017.

\bibitem[Higgins et~al.(2018)Higgins, Sonnerat, Matthey, Pal, Burgess, Bosnjak,
  Shanahan, Botvinick, Hassabis, and Lerchner]{higgins2017scan}
Irina Higgins, Nicolas Sonnerat, Loic Matthey, Arka Pal, Christopher~P Burgess,
  Matko Bosnjak, Murray Shanahan, Matthew Botvinick, Demis Hassabis, and
  Alexander Lerchner.
\newblock Scan: Learning hierarchical compositional visual concepts.
\newblock \emph{ICLR}, 2018.

\bibitem[Hinton(1999)]{hinton1999products}
Geoffrey~E Hinton.
\newblock Products of experts.
\newblock \emph{International Conference on Artificial Neural Networks}, 1999.

\bibitem[Hinton(2002)]{hinton2002training}
Geoffrey~E Hinton.
\newblock Training products of experts by minimizing contrastive divergence.
\newblock \emph{Neural computation}, 14\penalty0 (8):\penalty0 1771--1800,
  2002.

\bibitem[Hinton et~al.(2006)Hinton, Osindero, and Teh]{hinton2006fast}
Geoffrey~E Hinton, Simon Osindero, and Yee-Whye Teh.
\newblock A fast learning algorithm for deep belief nets.
\newblock \emph{Neural Comput.}, 18\penalty0 (7):\penalty0 1527--1554, 2006.

\bibitem[Kim and Bengio(2016)]{kim2016deep}
Taesup Kim and Yoshua Bengio.
\newblock Deep directed generative models with energy-based probability
  estimation.
\newblock \emph{arXiv preprint arXiv:1606.03439}, 2016.

\bibitem[Kirkpatrick et~al.(2017)Kirkpatrick, Pascanu, Rabinowitz, Veness,
  Desjardins, Rusu, Milan, Quan, Ramalho, Grabska-Barwinska,
  et~al.]{kirkpatrick2017overcoming}
James Kirkpatrick, Razvan Pascanu, Neil Rabinowitz, Joel Veness, Guillaume
  Desjardins, Andrei~A Rusu, Kieran Milan, John Quan, Tiago Ramalho, Agnieszka
  Grabska-Barwinska, et~al.
\newblock Overcoming catastrophic forgetting in neural networks.
\newblock \emph{Proceedings of the national academy of sciences}, 114\penalty0
  (13):\penalty0 3521--3526, 2017.

\bibitem[Kulkarni et~al.(2015)Kulkarni, Whitney, Kohli, and
  Tenenbaum]{Kulkarni2015Deep}
Tejas~D Kulkarni, William~F Whitney, Pushmeet Kohli, and Josh Tenenbaum.
\newblock Deep convolutional inverse graphics network.
\newblock In \emph{NIPS}, 2015.

\bibitem[Lake et~al.(2017)Lake, Ullman, Tenenbaum, and
  Gershman]{lake2017building}
Brenden~M Lake, Tomer~D Ullman, Joshua~B Tenenbaum, and Samuel~J Gershman.
\newblock Building machines that learn and think like people.
\newblock \emph{Behavioral and brain sciences}, 40, 2017.

\bibitem[LeCun et~al.(2006)LeCun, Chopra, and Hadsell]{lecun2006tutorial}
Yann LeCun, Sumit Chopra, and Raia Hadsell.
\newblock A tutorial on energy-based learning.
\newblock 2006.

\bibitem[Li and Hoiem(2017)]{li2017learning}
Zhizhong Li and Derek Hoiem.
\newblock Learning without forgetting.
\newblock \emph{IEEE transactions on pattern analysis and machine
  intelligence}, 40\penalty0 (12):\penalty0 2935--2947, 2017.

\bibitem[Mnih and Hinton(2005)]{mnih2005learning}
Andriy Mnih and Geoffrey Hinton.
\newblock Learning nonlinear constraints with contrastive backpropagation.
\newblock In \emph{Proceedings. 2005 IEEE International Joint Conference on
  Neural Networks, 2005.}, volume~2, pages 1302--1307. IEEE, 2005.

\bibitem[Mokady et~al.(2018)Mokady, Benaim, Wolf, and Bermano]{abs-1906.06558}
Ron Mokady, Sagie Benaim, Lior Wolf, and Amit Bermano.
\newblock Mask based unsupervised content transfer.
\newblock \emph{CoRR}, abs/1906.06558, 2018.
\newblock URL \url{http://arxiv.org/abs/1906.06558}.

\bibitem[Oord et~al.(2016)Oord, Kalchbrenner, and Kavukcuoglu]{oord2016pixel}
Aaron van~den Oord, Nal Kalchbrenner, and Koray Kavukcuoglu.
\newblock Pixel recurrent neural networks.
\newblock \emph{arXiv preprint arXiv:1601.06759}, 2016.

\bibitem[Parisi et~al.(2018)Parisi, Kemker, Part, Kanan, and Wermter]{parisi}
German~Ignacio Parisi, Ronald Kemker, Jose~L. Part, Christopher Kanan, and
  Stefan Wermter.
\newblock Continual lifelong learning with neural networks: {A} review.
\newblock \emph{CoRR}, abs/1802.07569, 2018.
\newblock URL \url{http://arxiv.org/abs/1802.07569}.

\bibitem[Press et~al.(2019)Press, Galanti, Benaim, and Wolf]{press2018emerging}
Ori Press, Tomer Galanti, Sagie Benaim, and Lior Wolf.
\newblock Emerging disentanglement in auto-encoder based unsupervised image
  content transfer.
\newblock In \emph{International Conference on Learning Representations}, 2019.
\newblock URL \url{https://openreview.net/forum?id=BylE1205Fm}.

\bibitem[Radford et~al.(2015)Radford, Metz, and
  Chintala]{radford2015unsupervised}
Alec Radford, Luke Metz, and Soumith Chintala.
\newblock Unsupervised representation learning with deep convolutional
  generative adversarial networks.
\newblock \emph{arXiv preprint arXiv:1511.06434}, 2015.

\bibitem[Ramachandran et~al.(2017)Ramachandran, Zoph, and
  Le]{ramachandran2017searching}
Prajit Ramachandran, Barret Zoph, and Quoc~V Le.
\newblock Searching for activation functions.
\newblock \emph{arXiv preprint arXiv:1710.05941}, 2017.

\bibitem[Reed et~al.(2017)Reed, Chen, Paine, Oord, Eslami, Rezende, Vinyals,
  and de~Freitas]{reed2017few}
Scott Reed, Yutian Chen, Thomas Paine, A{\"a}ron van~den Oord, SM~Eslami,
  Danilo Rezende, Oriol Vinyals, and Nando de~Freitas.
\newblock Few-shot autoregressive density estimation: Towards learning to learn
  distributions.
\newblock \emph{arXiv preprint arXiv:1710.10304}, 2017.

\bibitem[Rusu et~al.(2016)Rusu, Rabinowitz, Desjardins, Soyer, Kirkpatrick,
  Kavukcuoglu, Pascanu, and Hadsell]{rusu2016progressive}
Andrei~A Rusu, Neil~C Rabinowitz, Guillaume Desjardins, Hubert Soyer, James
  Kirkpatrick, Koray Kavukcuoglu, Razvan Pascanu, and Raia Hadsell.
\newblock Progressive neural networks.
\newblock \emph{arXiv preprint arXiv:1606.04671}, 2016.

\bibitem[Salimans et~al.(2017)Salimans, Karpathy, Chen, and
  Kingma]{salimans2017pixelcnn++}
Tim Salimans, Andrej Karpathy, Xi~Chen, and Diederik~P Kingma.
\newblock Pixelcnn++: Improving the pixelcnn with discretized logistic mixture
  likelihood and other modifications.
\newblock \emph{arXiv preprint arXiv:1701.05517}, 2017.

\bibitem[Shazeer et~al.(2017)Shazeer, Mirhoseini, Maziarz, Davis, Le, Hinton,
  and Dean]{shazeer2017outrageously}
Noam Shazeer, Azalia Mirhoseini, Krzysztof Maziarz, Andy Davis, Quoc Le,
  Geoffrey Hinton, and Jeff Dean.
\newblock Outrageously large neural networks: The sparsely-gated
  mixture-of-experts layer.
\newblock \emph{arXiv preprint arXiv:1701.06538}, 2017.

\bibitem[Song and Ou(2018)]{song2018learning}
Yunfu Song and Zhijian Ou.
\newblock Learning neural random fields with inclusive auxiliary generators.
\newblock \emph{arXiv preprint arXiv:1806.00271}, 2018.

\bibitem[Todorov et~al.(2012)Todorov, Erez, and Tassa]{Todorov2012MuJoCo}
Emanuel Todorov, Tom Erez, and Yuval Tassa.
\newblock Mujoco: A physics engine for model-based control.
\newblock In \emph{2012 IEEE/RSJ International Conference on Intelligent Robots
  and Systems}, pages 5026--5033. IEEE, 2012.

\bibitem[van Steenkiste et~al.(2018)van Steenkiste, Kurach, and
  Gelly]{van2018case}
Sjoerd van Steenkiste, Karol Kurach, and Sylvain Gelly.
\newblock A case for object compositionality in deep generative models of
  images.
\newblock \emph{arXiv preprint arXiv:1810.10340}, 2018.

\bibitem[Vedantam et~al.(2018)Vedantam, Fischer, Huang, and
  Murphy]{vedantam2017generative}
Ramakrishna Vedantam, Ian Fischer, Jonathan Huang, and Kevin Murphy.
\newblock Generative models of visually grounded imagination.
\newblock In \emph{ICLR}, 2018.

\bibitem[Welling and Teh(2011)]{welling2011bayesian}
Max Welling and Yee~W Teh.
\newblock Bayesian learning via stochastic gradient langevin dynamics.
\newblock In \emph{Proceedings of the 28th International Conference on Machine
  Learning (ICML-11)}, pages 681--688, 2011.

\bibitem[Xie et~al.(2016)Xie, Lu, Zhu, and Wu]{xie2016theory}
Jianwen Xie, Yang Lu, Song-Chun Zhu, and Yingnian Wu.
\newblock A theory of generative convnet.
\newblock In \emph{International Conference on Machine Learning}, pages
  2635--2644, 2016.

\end{thebibliography}
